%% file: main.tex
\documentclass[nohyperref]{article}

\usepackage{microtype}
\usepackage{graphicx}
\usepackage{subfigure}
\usepackage{booktabs} % for professional tables

\usepackage{hyperref}

\usepackage[accepted]{icml2023}

\usepackage{amsmath}
\usepackage{amssymb}
\usepackage{mathtools}
\usepackage{amsthm}
\usepackage{booktabs}       
\usepackage{amsfonts}       
\usepackage{nicefrac}       
\usepackage{microtype}   
\usepackage{xcolor}    
\usepackage{array,algorithm,etoolbox}
\usepackage{algorithmic}
\usepackage{graphicx}
\usepackage{subfigure}
\usepackage{textcomp}
\usepackage{footnote}
\usepackage{tablefootnote}
\usepackage{comment}
\usepackage{tikz}
\usepackage{multirow}

\newcommand{\cO}{\mathcal{O}}

\newcommand{\bE}{\mathbb{E}}
\newcommand{\bP}{\mathbb{P}}
\newcommand{\bI}{\mathbb{I}}
\newcommand{\PTP}{K^{1/3}T^{2/3}} % short for phase transition point
\newcommand{\wtTheta}{\widetilde{\Theta}}
\newcommand{\wtOmega}{\widetilde{\Omega}}
\newcommand{\wtO}{\widetilde{O}}
\newcommand{\Bex}{B_{\mathrm{ex}}}

\usepackage[toc,page,header]{appendix}
\usepackage{minitoc}

\newcommand{\secref}[1]{Section~\ref{sec:#1}}

\newcommand{\eqnref}[1]{Eq.~(\ref{eqn:#1})}
\newcommand{\thmref}[1]{Theorem~\ref{thm:#1}}

\newcommand{\algref}[1]{Algorithm~\ref{alg:#1}}

\newcommand{\lemref}[1]{Lemma~\ref{lem:#1}}

\newcommand{\ex}[1]{\mathbb{E}\left[#1\right]}

\newcommand{\duo}[1] {{\footnotesize\color{orange}[Duo: #1]}}
\newcommand{\bo}[1] {{\footnotesize\color{teal}[Bo: #1]}}

\usepackage[capitalize,noabbrev]{cleveref}

\theoremstyle{plain}
\newtheorem{theorem}{Theorem}
\newtheorem{proposition}{Proposition}
\newtheorem{lemma}{Lemma}

\theoremstyle{definition}

\theoremstyle{remark}
\newtheorem{remark}{Remark}

\newcommand{\alglinelabel}{%
  \addtocounter{ALC@line}{-1}% Reduce line counter by 1
  \refstepcounter{ALC@line}% Increment line counter with reference capability
  \label% Regular \label
}

\newcommand{\namedtheoremname}{}
\newtheorem{namedtheoreminner}[theorem]{\protect\namedtheoremname}

\usepackage[textsize=tiny]{todonotes}

\definecolor{DarkGreen}{rgb}{0.2,0.6,0.2}
	\definecolor{DarkRed}{rgb}{0.6,0.2,0.2}
	\definecolor{DarkBlue}{rgb}{0.2,0.2,0.6}
	\definecolor{DarkPurple}{rgb}{0.4,0.2,0.4}   
\hypersetup{
        linktocpage=true,
        colorlinks=true,				
        linkcolor=DarkBlue,				
        citecolor=DarkBlue,				
        urlcolor=DarkBlue,			
    }

\icmltitlerunning{Understanding the Role of Feedback in Online Learning with Switching Costs}

\begin{document}

% \input{notes.tex}
% \newpage

\twocolumn[
\icmltitle{Understanding the Role of Feedback in Online Learning with Switching Costs}



\icmlsetsymbol{equal}{*}

\begin{icmlauthorlist}
\icmlauthor{Duo Cheng}{a}
\icmlauthor{Xingyu Zhou}{b}
\icmlauthor{Bo Ji}{a}

\end{icmlauthorlist}

\icmlaffiliation{a}{Virginia Tech, Blacksburg, USA}
\icmlaffiliation{b}{Wayne State University, Detroit, USA}

\icmlcorrespondingauthor{Bo Ji}{boji@vt.edu}
%\icmlcorrespondingauthor{Firstname2 Lastname2}{first2.last2@www.uk}

\icmlkeywords{Machine Learning, ICML}

\vskip 0.3in
]



\printAffiliationsAndNotice{}  % leave blank if no need to mention equal contribution

\begin{abstract}
In this paper, we study the role of feedback in online learning with switching costs. It has been shown that the minimax regret is $\widetilde{\Theta}(T^{2/3})$ under bandit feedback and improves to $\widetilde{\Theta}(\sqrt{T})$ under full-information feedback, where $T$ is the length of the time horizon. However, it remains largely unknown how the amount and type of feedback generally impact regret. 
To this end, we first consider the setting of bandit learning with extra observations; that is, in addition to the typical bandit feedback, the learner can freely make a total of $B_{\mathrm{ex}}$ \emph{extra observations}. We fully characterize the minimax regret in this setting, which exhibits an interesting \emph{phase-transition phenomenon}: when $B_{\mathrm{ex}} = O(T^{2/3})$, the regret remains $\widetilde{\Theta}(T^{2/3})$, but when $B_{\mathrm{ex}} = \Omega(T^{2/3})$, it becomes $\widetilde{\Theta}(T/\sqrt{\Bex})$, which improves as the budget $B_{\mathrm{ex}}$ increases. To design algorithms that can achieve the minimax regret, it is instructive to consider a more general setting where the learner has a budget of $B$ \emph{total observations}. 
We fully characterize the minimax regret in this setting as well and show that it is $\wtTheta(T/\sqrt{B})$, which scales smoothly with the total budget $B$. 
Furthermore, we propose a generic algorithmic framework, which enables us to design different learning algorithms that can achieve matching upper bounds for both settings based on the amount and type of feedback. One interesting finding is that while bandit feedback can still guarantee optimal regret when the budget is relatively limited, it no longer suffices to achieve optimal regret when the budget is relatively large.

%\xingyu{finished it. not sure the last sentence is correct or not:). I think roughly it is correct.}\bo{Thanks, Xingyu. I will take a look now.}\xingyu{great}

% We provide optimal regret bounds by proving fundamental lower bounds and designing algorithms that can achieve them. These bounds unveil an interesting \emph{phase-transition phenomenon}.
% Specifically, when $\Bex = \wtO(\PTP)$, the regret cannot be improved and remains $\wtTheta(\PTP)$, but when $\Bex = \wtOmega(\PTP)$, one can exploit such extra observations to improve the regret to $\wtTheta(T\sqrt{K/\Bex})$. 
% In a more general setting of online learning with $B$ \emph{total observations}, we also derive the optimal regret bound of $\wtTheta( T\sqrt{{K}/{B}})$, which scales smoothly with the budget $B$. To achieve optimal regret in both settings, however, one needs to carefully choose feedback (e.g., bandit vs. full-information), especially when the budget is large (i.e., when $\Bex = \wtOmega(\PTP)$ or $B=\Omega(\PTP)$). Our findings reveal that both the amount and type of feedback play important roles in online learning with switching costs.\xingyu{I will give it a try first...}
\end{abstract}

\input{intro.tex}

\input{problem.tex}

% \input{sec4.tex}
\input{sec4-xingyu.tex}
\input{sec5-xingyu.tex}

\input{related.tex}
\section{Conclusion}
Our work is motivated by a well-known gap in the minimax regret under bandit feedback and full-information feedback in online learning with switching costs. We attempted to fundamentally understand the role of feedback by studying two cases of observation budget: (i) bandit feedback plus an extra observation budget and (ii) a total observation budget. Our findings reveal that both the amount and type of feedback play crucial roles when there are switching costs.
%We hope that this work would shed a light on the line of research towards a better understanding of how feedback plays a role in various other online learning problems. \xingyu{e.g.,?}

One interesting future direction is to consider stronger high-probability regret guarantees~\cite{neu2015explore}.
%, since our upper bounds hold in expectation only.
Another direction is to achieve \emph{the best of both worlds} guarantees for regrets with switching costs~\citep{Rouyer2021AnAF, amir2022better}.

\section*{Acknowledgments}
%\duo{use paragraph rather then sec then we are fine}\bo{you can use the original format, and we can make the page limit right later}
We thank the anonymous paper reviewers for their insightful feedback. This work is supported in part by the NSF grants under CNS-2112694 and
CNS-2153220.

\bibliography{ref}
\bibliographystyle{icml2023}
\input{app.tex}

\end{document}

%% file: intro.tex
\section{Introduction}
\label{sec:intro}
%\bo{You use both ``learner'' and ``learner''; choose one and use it consistently. Same comment for ``action'' and ``action''.}\duo{fixed}
Online learning over a finite set of actions is a classical problem in machine learning research. It can be formulated as a $T$-round repeated game between a learner and an adversary: at each round, the learner chooses one of the $K$ actions and suffers the loss of this chosen action, where the loss is determined by the adversary. At the end of each round, the learner receives some feedback and uses it to update her policy at the next round. The goal of the learner is to minimize the \emph{regret}, defined as the difference between her cumulative loss and that of the best fixed action in hindsight.

In terms of the type of feedback, two important settings have been extensively studied in the literature: bandit and full information. At each round, if the learner observes only the loss of the chosen action, then it is called \emph{bandit} feedback, and the game is called \emph{adversarial multi-armed bandits} (MAB) or \emph{non-stochastic bandits with adversarial losses} \citep{auer2002nonstochastic}. 
%\bo{add a reference?}\duo{done}. 
On the other hand,
if the losses of all $K$ actions are revealed to the learner, then it is called \emph{full-information} feedback, and the game becomes \emph{prediction with expert advice} \citep{cesa2006prediction}. %\bo{add a reference?}\duo{done}.

The regret in these two settings has been well understood. Specifically, the minimax regret is $\Theta(\sqrt{TK})$\footnote{We use standard big $O$ notations (e.g., $O$, $\Omega$, and $\Theta$); those with tilde (e.g., $\wtO$, $\wtOmega$, and $\wtTheta$) hide poly-logarithmic factors.} under bandit feedback \cite{auer2002nonstochastic,Audibert2009MinimaxPF} and is
$\Theta(\sqrt{T\ln{K}})$ under full information~\citep[Theorems~2.2 and 3.7]{cesa2006prediction}~\citep{hazan2016introduction}~\citep[Section 6.8]{orabona2019modern}. These results imply that learning under bandit feedback is \emph{slightly harder} than under full information, in the sense that the dependency on $K$ is worse ($\Theta(\sqrt{K})$ vs. $\Theta(\sqrt{\ln{K}})$). However, the scaling with respect to $T$ remains the same (i.e., $\Theta(\sqrt{T})$).

In the above standard settings, the learner is allowed to \emph{arbitrarily} switch actions at two consecutive rounds. However, in many real-world decision-making problems, switching actions may incur a cost (e.g., due to system reconfiguration and resource reallocation) \citep{zhang2005traffic, kaplan2011power}. %\bo{add a reference?}\duo{done}. 
Motivated by this practical consideration, a new setting called \emph{online learning with switching costs} has also been extensively studied \citep{Dekel2012OnlineBL, cesa2013online}. In this setting, the learner needs to pay an additional unit loss whenever she switches actions.
%\xingyu{do we need to cite the first work on this setting?}\duo{I feel the first work is Arora's policy regret, which however didn't explicitly target at switching costs. I would say the first one work really from the switching cost viewpoint is  \cite{cesa2013online}, which showed the $2/3$ lower bound but with drifting and unbounded losses.} \bo{we can cite both the first work and the most important work}
%whenever she plays an action different from the one in the previous round.

Interestingly, it has been shown that in this new setting, learning under bandit feedback is \emph{significantly harder} than under full information. Under full-information feedback, even with switching costs, the minimax regret remains $\Theta(\sqrt{T\ln{K}})$, which can be achieved by several algorithms such as Shrinking Dartboard (SD) \cite{geulen2010regret} and Follow-the-Perturbed-Leader (FTPL) \cite{devroye2013prediction}. On the other hand, \citet{dekel2014bandits} shows a (worse) lower bound of $\wtOmega(\PTP)$ for the bandit setting, which can be matched (up to poly-logarithmic factors) by the batched EXP3 algorithm \cite{Dekel2012OnlineBL}. These results reveal that introducing switching costs makes bandit problems \emph{strictly harder} than expert problems due to the worse dependency on $T$ (i.e., $\wtTheta(T^{2/3})$ vs. $\wtTheta(\sqrt{T})$).

%\xingyu{finish the bullets.}

\textbf{Our Contributions.}
While these two special cases have been well studied, it remains largely unknown how feedback impacts regret in general. To close this important gap, we aim to fundamentally understand the role of feedback (in terms of both amount and type) in online learning with switching costs. 
Our main contributions are as follows. %To the best of our knowledge, this is the first work that makes such an attempt.
%In particular, we take a two-step principled approach and summarize our main contributions as follows. 

%\bo{revisit the bullets}

\begin{figure}[!t] 
\centering     
\subfigure[Bandit feedback plus $\Bex$ extra observations] { \label{fig:case 1}      \includegraphics[width=0.85\columnwidth]{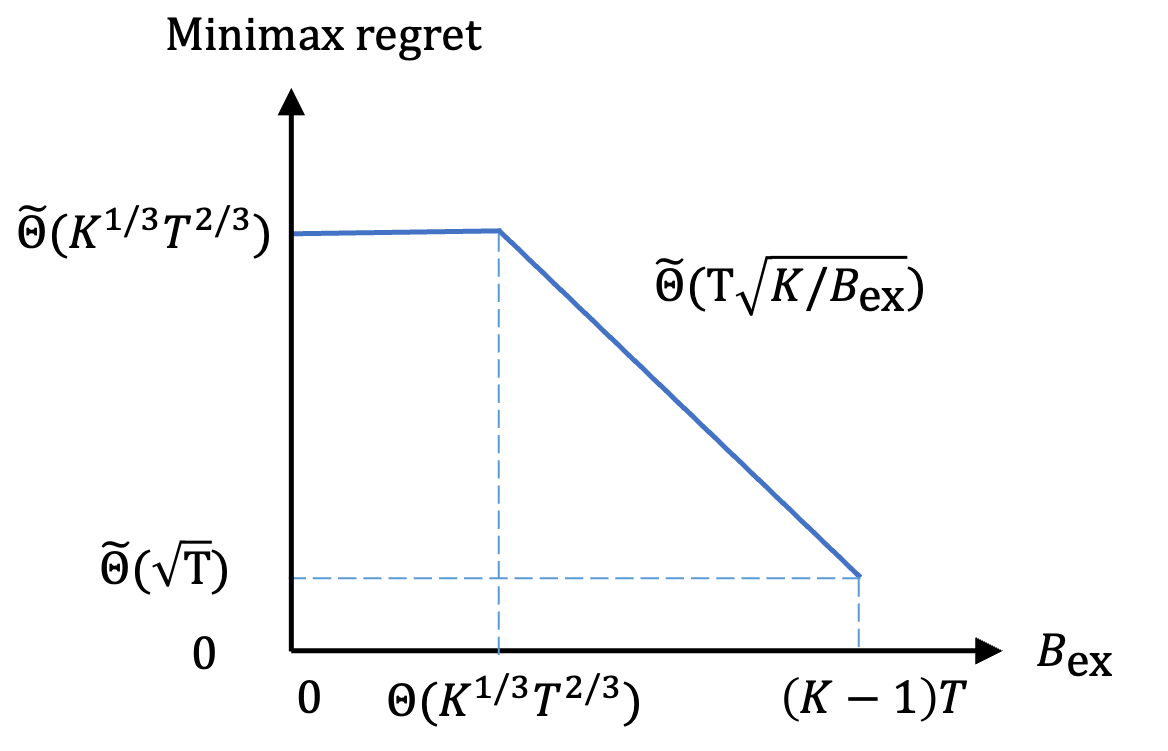}   }
\subfigure[$B$ total observations] { \label{fig:case 2}      \includegraphics[width=0.85\columnwidth]{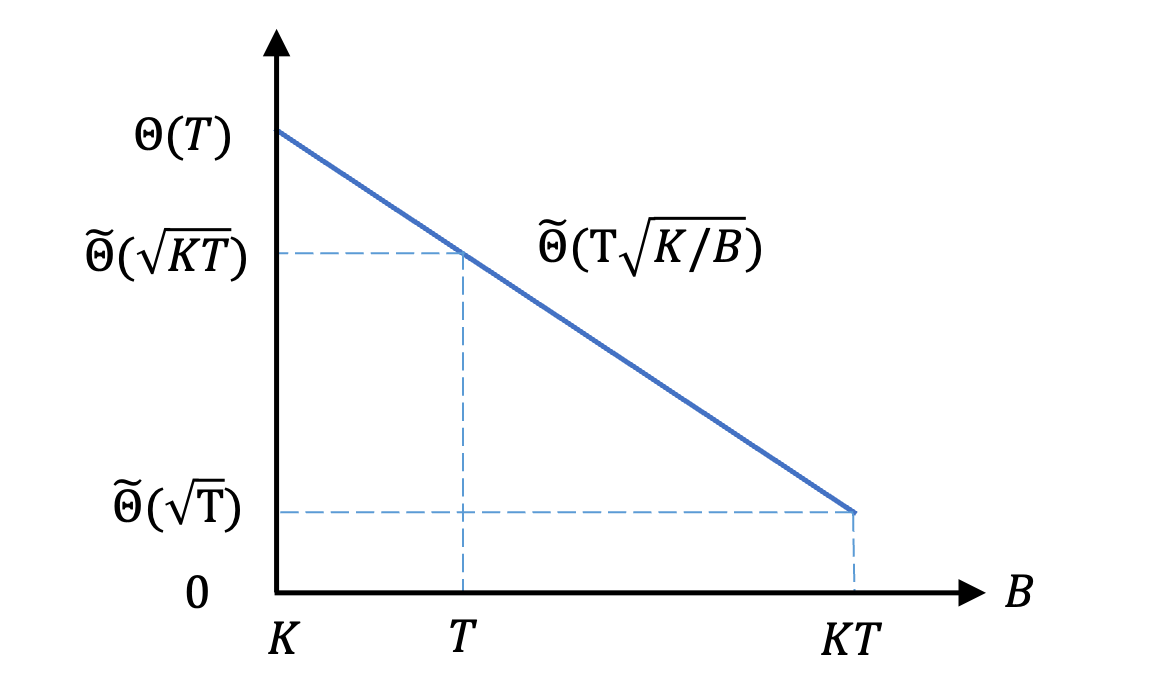}}      
\caption{ An illustration of the minimax regret vs. observation budget in log-log plots: (a) the learner receives bandit feedback plus no more than $\Bex$ extra observations (Theorem~\ref{thm:bandit_extra_minimax}); (b) the learner can make no more than $B$ total observations (Theorem~\ref{thm:total_minimax}). 
\vspace{-3mm}
}
\label{fig:main_results}      
\end{figure}

\textbf{(i)} We first consider the setting of bandit learning with extra observations, where in addition to the typical bandit feedback, the learner can freely make a total of $\Bex$ \emph{extra observations} in an arbitrary form (\cref{sec:extra}). We present a tight characterization of the minimax regret, which exhibits an interesting \emph{phase-transition phenomenon} (see Fig.~\ref{fig:case 1}). Specifically, when $\Bex = O(T^{2/3})$, the regret remains $\wtTheta(T^{2/3})$, but when $\Bex = \Omega(T^{2/3})$, it becomes $\wtTheta(T/\sqrt{\Bex})$, which improves as the budget $\Bex$ increases.
%To understand this, we draw useful insights from the lower bound analysis, which in turn motivates us to ask several fundamental questions on algorithm design, e.g., how to judiciously use extra feedback to match the lower bound?
%\bo{I am still not quite comfortable with this part. Why this question alone?}\xingyu{we can add some.}\duo{do we have to raise questions here?}\bo{I don't think so}\duo{then can we simply say ``To understand this, we draw useful insights from the lower bound analysis, which in turn motivates us in the algorithm design''?}
% \bo{we said questions, but there is only one}\duo{``algorithm design, e.g., how to \dots''?}

\textbf{(ii)} To understand this phenomenon and design algorithms that can achieve the minimax regret, it is instructive to consider a more general setting where the learner has a budget of $B$ \emph{total observations} (\cref{sec:total}). 
We fully characterize the minimax regret in this setting as well and show that it is $\wtTheta(T/\sqrt{B})$, which scales smoothly with the total budget $B$ (see Fig.~\ref{fig:case 2}). 
Furthermore, we propose a generic algorithmic framework, which enables us to design different learning algorithms that can achieve matching upper bounds for both settings based on the amount and type of feedback.

%\textbf{(ii)} To answer these questions, it turns out that it would be more convenient to consider a general setting where the learner has a budget for \emph{total} observations (\cref{sec:total}). For this new setting, we also present a tight characterization of the minimax regret, which now scales smoothly with respect to the total budget $B$ (see Fig.~\ref{fig:case 2}). We then propose a generic algorithmic framework, which allows us to design different learning algorithms based on the amount and type of feedback. More importantly, not only can these algorithms match the lower bound for this new setting, but they also help answer the fundamental questions we raised for the previous setting, hence completing our study.

\textbf{(iii)} Our findings highlight the crucial impact of feedback type (bandit vs. others) in the second setting (see Table~\ref{tab:regret_of_different_types}).
In particular, while both bandit and other types of feedback can achieve optimal regret when the budget is relatively limited, \emph{pure bandit feedback is no longer sufficient to guarantee optimal regret when the budget is relatively large.}
However, in the standard setting without switching costs, all three types of feedback we consider can achieve optimal regret in the full range of $B$.
This reveals that the impact of feedback type is (partly) due to switching costs. 

% \begin{table}[!t]
% \centering
% \caption{The minimax regret under different types of feedback in the setting of online learning with switching costs under a total observation budget $B$. \duo{``Flexible'' feedback is formally specified by $\pi_{\mathrm{flex}}$ in Proposition \ref{prop:flex upper bound}.} \bo{``Flexible'' is unclear here}\duo{add sth. in the caption?}}
% %\bo{need to refer back to this table in Section~4}}
% \vspace{2mm}
% \begin{tabular}{|c|c|}
% \hline
% \textbf{Feedback Type}                                                & \textbf{Minimax Regret} \\ \hline
% Full-information &
%   \multirow{3}{*}{\begin{tabular}[c]{@{}c@{}}$\wtTheta(T\sqrt{K/B})$\\ \!\!\! Lemma~\ref{lem:total_lower} + Propositions~\ref{prop:uniform_alg_upper_bound}, \ref{prop:flex upper bound}, \ref{prop:bandit_upper_bound}\end{tabular}}\!\!\! \\ \cline{1-1}
% Flexible                                                       &                \\ \cline{1-1}
% \begin{tabular}[c]{@{}c@{}}Bandit\\ $B=O(\PTP)$\end{tabular} &                \\  \hline
% \\[-1.1em]
% \begin{tabular}[c]{@{}c@{}}Bandit\\ $B=\Omega(\PTP)$\end{tabular} &
%   \begin{tabular}[c]{@{}c@{}}$\wtTheta(\PTP)$ \\ \!\!\!\citet{dekel2014bandits} + Proposition~\ref{prop:bandit_upper_bound}\end{tabular}\!\!\! \\ \hline
% \end{tabular}
% \end{table}

\begin{table}[!t]
\caption{The minimax regret under different types of feedback in the setting of online learning under a total observation budget $B$: with (w/) vs. without (w/o) switching costs (SC). A formal description of ``Flexible'' feedback can be found in Section~\ref{sec:full and flex alg}.}
\vspace{2mm}
\begin{tabular}{|c|cc|}
\hline
\multirow{2}{*}{\begin{tabular}[c]{@{}c@{}}\textbf{Feedback Type}\end{tabular}} & \multicolumn{2}{c|}{\textbf{Minimax Regret}} \\ \cline{2-3} 
 & \multicolumn{1}{c|}{w/ SC} &~~~~ w/o SC~~~~ \\ \hline
Full-information & \multicolumn{2}{c|}{\multirow{3}{*}{\vspace{-4mm}$\wtTheta(T\sqrt{K/B})$}} \\ \cline{1-1}
Flexible & \multicolumn{2}{c|}{} \\ \cline{1-1}
\begin{tabular}[c]{@{}c@{}}Bandit\\ ($B=O(\PTP)$)\end{tabular} & \multicolumn{2}{c|}{} \\ \cline{1-2}
\begin{tabular}[c]{@{}c@{}}Bandit\\ ($B=\Omega(\PTP)$)\end{tabular} & \multicolumn{1}{c|}{$\wtTheta(\PTP)$} &  \\ \hline
\end{tabular}\label{tab:regret_of_different_types}
\end{table}

%% file: problem.tex
\section{Problem Setup  }\label{sec:problem setup}
%\bo{Problem Setup instead of Problem
In this section, we introduce basic notations and present the problem setup. For any positive integer $n$, let $[n]:=\{1, \dots, n\}$, and let $\ell_{1:n}$ be the loss sequence $\ell_{1}, \dots, \ell_{n}$.
%\bo{use dots (instead of cdots), which will decide cdots or ldots automatically}
We use $\bI_{\{\mathcal{E}\}}$ to denote the indicator function of event $\mathcal{E}$: $\bI_{\{\mathcal{E}\}}=1$ if event $\mathcal{E}$ happens, and $\bI_{\{\mathcal{E}\}}=0$ otherwise.
%\bo{You later use $N$ for number of batches. I'd suggest that you design your notation system properly and keep a list of notations for yourself, even though you may not include it in the paper.}

The learning problem can be viewed as a repeated game between a learner and an adversary. Assume that there are $K > 1$ actions the learner can choose. Let $T \ge K$ be the length of the time horizon, which is fixed at the beginning of the game and is known to the learner. At each round $t \in [T]$, the adversary assigns a loss in $[0, 1]$ to each action in $[K]$; the learner samples an action $X_t$ from a probability distribution $w_t$ (also determined by the learner) over the action set $[K]$.
%\bo{You need to explain what an oblivious adversary is and why we consider randomized algorithms}\duo{1. will do. 2. I thought deterministic algs are just special cases when you put all the mass in single action. I'll see how other papers do regarding this.}\bo{If you recall our discussion in the class, the adversary can simulate the entire process of a deterministic algorithm and thus construct loss functions that are completely against the learner.}\duo{yes, I understand that deterministic learner is not able to learn. But since they are included by randomized players, we are WLOG. That's my thought. Anyway I'll see how other papers do regarding this.} 
After taking action $X_t$, the learner suffers a loss of the chosen action, i.e., $\ell_t[X_t]$. By the end of each round, the learner observes the loss of some actions (specific types of such feedback will be discussed later) and updates probability distribution $w_{t+1}$ that will be used at the next round.
%\xingyu{for next action?}\duo{i mean ``decision''}
Each time when the learner takes an action different from that at the previous round, one unit of switching cost is incurred. The \emph{regret} under a learning algorithm $\pi$ over a loss sequence $\ell_{1:T}$, denoted by $R^{\pi}_T(\ell_{1:T})$, is defined as the difference between the cumulative loss (including the switching costs incurred) under algorithm $\pi$ and that of the optimal (best fixed) action in hindsight:
%\bo{you use $x$ to denote an action in the following definition, but it looks like you use $k$ later. Be consistent.}
\begin{equation}\label{eqn:reg_def}
\small R^{\pi}_T(\ell_{1:T}) \!:=\!  \sum_{t=1}^{T}\left(\ell_{t}[X_{t}] \!+\! \bI_{\{X_{t} \neq X_{t-1}\}}\right) \!-\!\min _{k \in[K]} \sum_{t=1}^{T} \ell_{t}[k].
\end{equation}
For a randomized algorithm, we consider the \emph{expected regret} (or simply regret), denoted by $\ex{R^{\pi}_T(\ell_{1:T})}$, where the expectation is taken over the randomness of the algorithm. 
%where the randomness is w.r.t. $\{X_t\}_{t=1}^T$.
Without loss of generality, let $\bI_{\{X_{1} \neq X_{0}\}} = 0$, i.e., the first action does not incur any switching cost.
%\duo{I'm now sure it's 0, not 1 :)}
The adversary is assumed to be \emph{oblivious}, in the sense that the whole loss sequence is determined by the adversary before the game begins. In this paper, for any given algorithm $\pi$, we are interested in the \emph{worst-case (expected) regret} over all possible loss sequences (i.e., instance-independent), denoted by $R_T^{\pi}$:
\begin{equation}
\label{eq:reg-def}
    R_T^{\pi}:= \sup_{\ell_{1:T}\in [0,1]^{KT}} \ex{R^{\pi}_T(\ell_{1:T})}.
\end{equation}
% \bo{Your current way of using $R_T$ and $R_T^{\pi}$ could be confusing}\duo{now I use $R_T$ only in the analysis of the lower bound in sec 4. is it ok? i.e., ``For notational ease, we use $\ex{R_T^{\mathrm{unclipped}}}$ and $\ex{R_T}$ to denote them respectively in the rest of this section.'' other than that, there's no $R_T$ anywhere}
% \bo{Sounds good to me. Then, you do not even need to mention $R_T$ below Eq.~(1). Just introduce it in the lower bound proof.}\duo{yes. so I delete that sentence here}\bo{you still have it below Eq.~(1)}\duo{yes you are right. fixed. let me double check elsewhere}\bo{great.}
Let $\Pi$ be the set of all feasible learning algorithms following the specified learning protocol. %(For example, in bandit (expert) problems, $\Pi$ denotes all the learning algorithms relying on bandit (full) feedback.); 
We define the \emph{minimax (or optimal) regret}, denoted by $R_T^*(\Pi)$, as the minimum worst-case regret under all feasible learning algorithms in $\Pi$:
\begin{equation}\label{eqn:minimax_reg_def}
    R_T^*(\Pi) := \inf_{\pi\in\Pi} R_T^{\pi}.
\end{equation}
%that is, the smallest worst-case regret achieved by a learner from $\Pi$. 
For notational ease, we may drop $\Pi$ in $R_T^*(\Pi)$ and simply use $R_T^*$ whenever there is no ambiguity.
%\bo{I think we can use $R_T^*(\Pi)$ for simplicity.}\duo{we are already using $R_T^*(\Pi)$, right?}\bo{Sorry, I meant $R_T^*$}\duo{how about this. I don't think it's good to simplify in the definition.}\bo{That is fine.}
%Next, we state two assumptions.
%\bo{These assumptions are quite standard and do not need to be stated in the assumption form. We can merge them with the previous paragraph}\duo{done}

To understand the role of feedback in online learning with switching costs, we will consider two different settings with an observation budget: (i) in addition to the typical bandit feedback, the learner can freely make a total of $\Bex$ extra observations (Section~\ref{sec:extra}); (ii) the learner can freely make $B$ total observations (Section~\ref{sec:total}). 
Due to space limitations, in Appendix~\ref{app:motivate} we provide motivating examples for the settings with an observation budget we consider.
% \duo{For a motivating real-world example of learning with total observation budget, please see Appendix \ref{app:motivate} due to limited space.}
%\bo{choose one between ``observation budget'' and ``feedback budget'' and keep it consistent}\duo{now I make them all ``observation budget''}
%The details of these settings will be discussed later.
%both in which the learner is given some total budget such at she can freely make observations within it.

%\bo{We may want to discuss the feedback budget here and explain that we will consider two different cases.}\duo{how about this $\uparrow$}

%% file: sec4-xingyu.tex
\section{Bandit Learning with Switching Costs under Extra Observation Budget}\label{sec:extra}
%\duo{have changed the thm. statement into minimax, and mentioned the upper bound at the last line of sec 4. the title/division of subsec 4.3 4.4 may be weird}

Observing the gap in the optimal regret bound under bandit and full-information feedback ($\wtTheta(T^{2/3})$ vs. $\wtTheta(\sqrt{T})$),
%\footnote{In the rest of the paper, we will focus on the dependency on $T$ and may omit terms involving $K$ whenever the context is clear.} 
it is natural to ask: 
\emph{How much can one improve upon the $\wtTheta(T^{2/3})$ regret if the learner is allowed to make some extra observations in addition to the typical bandit feedback?} 
%In particular, if the learner can freely make no more than $\Bex$ extra observations throughout the game, what would be the optimal regret with respect to both $T$ and $\Bex$?

Motivated by this question, we consider the setting of bandit learning with switching costs under an \emph{extra} observation budget. We consider the learning protocol specified in \secref{problem setup}, and in addition to the typical bandit feedback, the learner is allowed to freely use at most $\Bex$ extra observations of the loss of other action(s) throughout the game, where $\Bex$ is an integer in $[0, (K-1)T]$. At the two endpoints of $0$ and $(K-1)T$, this new setting recovers the bandit and full-information cases, respectively. In this section, by slightly abusing the notation, we also use $\Pi$ to denote the set of all learning algorithms using typical bandit feedback plus $\Bex$ extra observations, and 
we are interested in the minimax regret $R^*_T$ for $\Bex \in [0, (K-1)T]$.
%\bo{you used this term, but never rigorously defined it}\duo{done. defined in eq. \ref{eqn:minimax_reg_def}}
% for general $\Bex$ in this range. Hence in this setting, $\Pi$ becomes the set of all learning algorithms relying on bandit feedback plus $\Bex$ extra observation.

\subsection{Minimax Regret}
We first present our main result of the minimax regret $R_T^*$ in this setting, which is formally stated in \thmref{bandit_extra_minimax}.
%To show this lower bound, we adopt the instance construction from \citet{shi2022power}, which is further a modification based on the multi-scale random walk proposed by \citet{dekel2014bandits}, with our refined analysis. \bo{The current way of writing may get the reader distracted to think about the difference between your work and \citet{shi2022power}. We should just say that our main idea is based on the multi-scale random walk here and mention \citet{shi2022power} later and explain the difference}\duo{can I keep the first sentence only? everything regarding the instance construction will be addressed in proof sketch}

%\bo{Use different counters for lemmas, theorems, remarks, etc.}\duo{done}
\begin{theorem}\label{thm:bandit_extra_minimax}
    In the setting of bandit learning with switching costs under an extra observation budget $\Bex \in [0, (K-1)T]$, the minimax regret is given by
    \begin{equation*}
        R_T^*=\left\{
        \begin{aligned}
        &\wtTheta(\PTP), & \Bex=O(\PTP), \\
        &\wtTheta(T\sqrt{K/\Bex}), & \Bex=\Omega(\PTP).
        \end{aligned}
        \right.
    \end{equation*}
\end{theorem}

%\bo{Since you focus on the minimax optimal regret $R_T^*$, I think you should state Theorem~1 based on a lower bound on the minimax optimal regret $R_T^*$. Then, you can probably use a big bracket with two cases, and no need to mention any learning algorithm and the existence of a loss sequence here in the statement (but only in the proof and/or proof sketch). Also check how other papers deal with this.}

%\bo{I think we can simply use $R^*_T$. In Section 5, we can say that by slightly abusing the notation, we also use $\Pi$ and $R^*_T$}\duo{for sure we can do that, but what's wrong this the current one which avoids further ambiguity?}\bo{I think this notation is unnecessarily complicated}

%\bo{Can we use the big bracket notation for two cases?}\duo{done}

%\bo{Similar comment for the upper bound.}\duo{I don't think we should mention minimax bound in the theorem statement of upper bound. However we can/should interpret it after the upper bound theorem}\bo{Why not? You can have a similar statement. In the proof sketch and proof, you can say that there exists an algorithm...}\duo{but we also need to introduce different algorithms}
%\bo{That is fine - it does not impact the theorem statement} \duo{we have theorem \ref{thm:general_upper_bound}, which is sufficient since it fully characterizes the optimal regret. after that we can focus on introducing the algorithm}
%\bo{That's what I meant by upper bound. For the propositions, of course, you need to introduce algorithms.}

%\bo{make this paragraph a remark?}\duo{done}

\begin{remark}
%\duo{I think there is a command manually shortening the vertical margins? Not sure if we can/need use}\bo{yes, vspace, but let's use this magic later ;-)}
Interestingly, this minimax regret exhibits a \emph{phase-transition phenomenon} (see, also, Fig.~\ref{fig:case 1}):
when the amount of extra observations is relatively small (i.e., $\Bex = O(\PTP)$), they are insufficient for improving the regret, which remains $\wtTheta(\PTP)$; however, when the amount is large enough (i.e., $\Bex = \Omega(\PTP)$), the regret decreases smoothly as the budget $\Bex$ increases.
\end{remark}

% \xingyu{this seems redundant.}\bo{Do you think a proposition on the lower bound is unnecessary? I felt that there was a gap between 4.1 and 4.2.}\xingyu{since the two results are basically the same? }

% To prove Theorem~\ref{thm:bandit_extra_minimax}, we need to show a fundamental lower bound and provide algorithms that can achieve this lower bound. Next, we present a lower bound of the minimax regret. The result is stated in the following proposition.

% \begin{proposition}\label{prop:bandit_extra_lower_bound}
%      In the setting of bandit learning with switching costs under an extra observation budget $\Bex \in [0, (K-1)T]$, the minimax regret has the following lower bound:
%     \begin{equation*}
%         R_T^*=\left\{
%         \begin{aligned}
%         &\wtOmega(\PTP), & ~\text{if}~\Bex=\wtO(\PTP), \\
%         &\wtOmega(T\sqrt{K/\Bex}), & ~\text{if}~\Bex=\wtOmega(\PTP).
%         \end{aligned}
%         \right.
%     \end{equation*}
% \end{proposition}

\subsection{Lower Bound}
To establish Theorem~\ref{thm:bandit_extra_minimax}, we will first show a fundamental lower bound, which is formally stated in Proposition~\ref{prop:bandit_extra_lower_bound}.

\begin{proposition}\label{prop:bandit_extra_lower_bound}
    For any learning algorithm $\pi$ that can use a total of $\Bex$ extra observations in addition to the typical bandit feedback, there exists a loss sequence $\ell_{1:T}$ (which may depend on both $\pi$ and $\Bex$) such that 
    \begin{equation*}
        \!\ex{R_T^{\pi}(\ell_{1:T})}\!=\!\left\{
        \begin{aligned}
        &\wtOmega(\PTP), \!\!\!\!&\!\Bex=O(\PTP), \\
        &\wtOmega(T\sqrt{K/\Bex}), \!\!\!\!&\!\Bex=\Omega(\PTP).
        \end{aligned}
        \right.
    \end{equation*}
\end{proposition}

% \begin{proposition}\label{prop:bandit_extra_lower_bound}
%      In the setting of bandit learning with switching costs under an extra observation budget $\Bex \in [0, (K-1)T]$, the minimax regret has the following lower bound:
%     \begin{equation*}
%         R_T^*=\left\{
%         \begin{aligned}
%         &\wtOmega(\PTP), & ~\text{if}~\Bex=\wtO(\PTP), \\
%         &\wtOmega(T\sqrt{K/\Bex}), & ~\text{if}~\Bex=\wtOmega(\PTP).
%         \end{aligned}
%         \right.
%     \end{equation*}
% \end{proposition}

% \subsection{Proof Sketch of the Lower Bound}
We provide detailed proof of the above lower bound in Appendix~\ref{app:extra_lower_bound_proof}. Here, we present a proof sketch that mainly focuses on the key steps of the lower bound analysis with necessary explanations. The proof sketch reveals useful insights that not only help explain the interesting phase-transition phenomenon but also shed light on the design of algorithms that can achieve this lower bound. 

\begin{proof}[Proof Sketch of Proposition~\ref{prop:bandit_extra_lower_bound}]

% After that, we will further show that some observations from this lower bound analysis indeed motivate our algorithm design 
% so that we can match this lower bound, which will be discussed later in this paper. 

% In the following, we explain the main idea of it along with a proof sketch and provide useful insights into the interesting phase-transition phenomenon. 

    % To prove this lower bound, it suffices to show that for any learning algorithm $\pi$ that can use a total of $\Bex$ extra observations in addition to bandit feedback, there exists a loss sequence $\ell_{1:T}$ (which may depend on both $\pi$ and $\Bex$) such that when $\Bex = \wtO(\PTP)$, we have $\ex{R_T^{\pi}(\ell_{1:T})} =  \wtOmega(\PTP)$; and when $\Bex = \wtOmega(\PTP)$, we have $\ex{R_T^{\pi}(\ell_{1:T})} =  \wtOmega( T\sqrt{K/\Bex})$.
    % \duo{how about making this statement as proposition 1? In this case we can avoid repeat theorem 1 twice (only Theta is replaced by Omega)}\xingyu{I agree. I tend to like to replace the current Prop 1, with the above sentences.}
    %\bo{Is the construction the same in two cases or not?}\duo{no. e.g., the choice of $\epsilon$ is different.}
    
    We first give an overview of the construction of hard loss sequences in our setting and the main ideas behind the construction.

    Generally speaking, the difficulty of bandit problems lies in the \emph{exploitation-exploration} tradeoff.
    %\bo{be consistent throughout the paper; I'd prefer to use ``tradeoff'' as it is now adopted as a single word; similar for nontrivial, nonnegative, etc.; check dictionary}
    On the one hand, the learner wants to pull empirically good actions in order to enjoy a low instantaneous loss (i.e., exploitation); on the other hand, she may also want to pull other actions and gain useful information to distinguish the optimal (best fixed) action and suboptimal actions (i.e., exploration).
    %gain more distinguishability on the identity of the underlying optimal action (i.e., exploration).

    In the presence of switching costs,
    %\duo{how about deleting: (one intuitive way to construct hard instances is to \duo{construct a relationship between} switches and exploration. \bo{``connect'' is unclear} In particular,)}
    \citet{dekel2014bandits} proposes hard instances (i.e., loss sequences) based on a \emph{multi-scale random walk} such that useful information toward distinguishability (between the optimal action and suboptimal actions) \emph{can only be obtained when the learner switches actions}, which, however, incurs switching costs. Using carefully constructed instances, they show that switching costs increase the intrinsic difficulty of bandit learning and result in a regret lower bound of $\wtOmega(\PTP)$. 
    %\bo{$\wtOmega(\PTP)$?}\duo{yes}

    However, the hard instances in \citet{dekel2014bandits} work for \emph{pure bandit feedback} only. That is, if the learner can obtain full-information feedback at \emph{any} of the $T$ rounds, she would immediately identify the optimal action and suffer no regret in the rest of the game. The reason is that the optimal action has the (unique) lowest loss at \emph{all} $T$ rounds.

    To make it still hard to learn even when the learner has some extra feedback, we will borrow an idea from \citet{shi2022power} to modify the original hard instance in~\citet{dekel2014bandits}: at each round, an additional layer of action-dependent noise is added to the loss of each action. As a result, the optimal action no longer has the lowest loss at all rounds and therefore cannot be trivially identified even when the learner can make extra observations.
    %beyond the typical bandit feedback. 

    In the rest of the proof sketch, we present three key steps of the proof and provide high-level explanations. 
    %\bo{what are the illustrations here?}\duo{by ``illustrations'' I meant ``explanations''}\bo{how about ``provide high-level explanations''}\duo{done}
    
    % In the rest of this proof sketch, we follow the organization of Appendix G.3 from \citet{altschuler2018online} to show key steps and deliver some illustrations.
    
    \textbf{Step 1: Establishing the relationship between two regrets.}
    %\bo{Here, it is unclear what kind of random walk you refer to and what the truncated version is used for. You need to explain it clearly.}
    As in~\citet{dekel2014bandits}, each loss value in the initial loss sequence we construct,
    %\bo{referring two papers here is a little confusing}\duo{only cite the first one?}\xingyu{yes, I prefer not to cite shi here}
    denoted by $\ell_{1:T}^{\mathrm{init}}$, may not be bounded in $[0,1]$; through truncation, we construct the actual loss sequence $\ell_{1:T}$ by simply projecting each initial loss value onto $[0,1]$.
    %$\ell_{1:T}^{\mathrm{init}}$ onto $[0,1]^T$
    %\duo{should be $[0,1]^{KT}$. Moreover, if readers view this as a high-dim projection, i'm not sure if it's confusing}\bo{how about now?}\duo{yes it's fine}.
    %\bo{Isn't this just projection to $[0,1]$? If so, just say it, and no need to have the footnote}\duo{I'm not sure if ``projection'' would be clear without a definition. If so then i'm fine}\bo{I think it should be fine, just say a simple project to $[0,1]$}
    For notational ease, we use $R_T^{\mathrm{init}}$ and $R_T$ to denote the regret over loss sequences $\ell_{1:T}^{\mathrm{init}}$ and $\ell_{1:T}$, respectively.
    Recall that the goal is to obtain a lower bound on $\ex{R_T}$, which, however, is hard to analyze directly due to the truncation. Instead, we show that it suffices to obtain a lower bound on $\ex{R_T^{\mathrm{init}}}$ (i.e., the regret under untruncated loss sequence), due to the following relationship:
    \begin{align}\label{eqn:unclipped_bounds_clipped}
        \ex{R_T} \geq \ex{R_T^{\mathrm{init}}} - \frac{\epsilon T}{6},
    \end{align}
    where $\epsilon>0$ is the gap between the instantaneous losses of the optimal action and a suboptimal action. The value of $\epsilon$ will be determined later. %\eqnref{unclipped_bounds_clipped} implies that a lower bound on $\ex{R_T^{\mathrm{init}}}$ also leads to a lower bound on $\ex{R_T}$. 

\textbf{Step 2: Obtaining a lower bound on $\ex{R_T^{\mathrm{init}}}$.}
%\duo{isn't this our ultimate goal? add sth. like ``in terms of the number of switches''?}}\bo{I think we should be fine. We mentioned this right above Eq.~(5).} 
Let $S$ be the expected total number of action switches. Through careful information-theoretic analysis, we obtain the following (informal) lower bound on $\ex{R_T^{\mathrm{init}}}$ in terms of the number of switches $S$ and extra observation budget $\Bex$:
\begin{align}
   \!\ex{R_T^{\mathrm{init}}} \!\ge\!
   \underbrace{\!\frac{\epsilon T}{2}}_{\mathbf{A.1}}\!-\!\underbrace{C \frac{ \epsilon^2 T }{\sqrt{K}} ( \sqrt{S} \!+\!  \sqrt{ \Bex})}_{\mathbf{A.2}} \!+\! \underbrace{S}_{\mathbf{A.3}}\label{eqn:extra_lower_bound_before_minimize},
\end{align}
% \begin{align}
%    \ex{R_T^{\mathrm{init}}} &\geq\nonumber \\
%    \underbrace{\!\frac{\epsilon T}{2}}_{\mathbf{A.1}}&\!-\!\underbrace{C \frac{ \epsilon^2 T }{\sqrt{K}} \left( \sqrt{ N^\mathrm{s}} \!+\!  \sqrt{ \Bex}\right)}_{\mathbf{A.2}} \!+\! \underbrace{S}_{\mathbf{A.3}}\label{eqn:extra_lower_bound_before_minimize},
% \end{align}
%where $S$
%\bo{how about $N_\mathrm{s}$?}\duo{this somehow conflicts some notations in the appendix \ref{app:extra_lower_bound_proof}}\bo{change $S^*$ accordingly} 
%is the total number of switches and 
where $C$ is a positive term that contains some constants and poly-logarithmic terms of $T$.
%\bo{use $S$ instead? Actually, why not simply $S$?}\duo{because $S$ has been used in theappendix for different meaning.} 
%\bo{Is it possible that you use $S$ here and use $\bar{N}^s$ where you currently use $S$?}\duo{done}\bo{Make sure to double check when you simply find and replace. It still has the bar in $\bar{N}_*^s$}\duo{I have iterated every ``bar'' command}
%Here, we choose $\sigma=\Theta(1/\log{T})$, which is contained in $C_1$. \bo{There is no $\sigma$. Delete this sentence?}\duo{Yes. I forgot about it}

%We now explain each term on the right-hand side (RHS) of \eqnref{extra_lower_bound_before_minimize}. $\mathbf{A.1}$ %\bo{remove the word ``Term'' in the equations?}\duo{done}is the regret that the learner will suffer without any distinguishability \bo{you use distinguishability a lot here; make sure you explain this clearly the first time you use this term}\duo{I explain it in the newly-added paragraph which introduces the ex-ex dilemma}, namely, $\Theta(\epsilon T)$ \bo{why is there $1/2$ in A.1?}. $\mathbf{A.2}$ corresponds to  how much information the learner collects that is useful for distinguishing the optimal 
%\bo{use ``optimal'' throughout the paper?}\duo{I can do that but it's a little weird to say ``optimal fixed action''. Ppl normally say ``optimal fixed'' }\duo{done} action from suboptimal actions: better distinguishability leads to a larger $\mathbf{A.2}$, which further leads to a smaller regret lower bound. $\mathbf{A.3}$ is simply the switching costs incurred.

We now explain each term in Eq.~\eqref{eqn:extra_lower_bound_before_minimize}.
%on the right-hand side (RHS) of \eqnref{extra_lower_bound_before_minimize} to characterize how the lower bound is established. 
Term~$\mathbf{A.1}$ reflects that without any useful information toward distinguishability, the learner may be stuck with a suboptimal action throughout the game, thus suffering $\Theta(\epsilon T)$ regret. Term~$\mathbf{A.2}$ roughly represents the amount of useful information for gaining distinguishability and thus reducing the regret: better distinguishability leads to a larger $\mathbf{A.2}$ and thus a lower regret. 
Term~$\mathbf{A.3}$ is simply the switching costs incurred.

\textbf{Step 3: Choosing a proper value of $\epsilon$.} 
%according to the value of $\Bex$.}
Note that the lower bound in \eqnref{extra_lower_bound_before_minimize} is a quadratic function of $\sqrt{S}$. By finding the minimizer of this quadratic function, denoted by $S^*$, we can further obtain the following lower bound:
\begin{align}
    \ex{R_T^{\mathrm{init}}} \geq
   \underbrace{\frac{\epsilon T}{2}}_{\mathbf{B.1}}-~\underbrace{\frac{C^2}{4}\cdot \frac{ \epsilon^4 T^2 }{K}}_{\mathbf{B.2}} - \underbrace{C\frac{\epsilon^2 T\sqrt{\Bex}}{\sqrt{K}}}_{\mathbf{B.3}}.
   \label{eqn:lower_bound_after_minimize}    
\end{align}
 %where $C_2$ and $C_3$ \bo{shouldn't we simply have $C_3=C_1$ and $C_2=C_1^2$? If so, just use $C$ instead of $C_1$}\duo{yes, fixed}\xingyu{no $C_2$ $C_3$ now} contain some universal constants and logarithmic terms of $T$.
 It now remains to choose a proper value of $\epsilon$ based on $\Bex$. 
 %\bo{Does this imply the construction is the same for these two cases?}\duo{in some sense yes. only the choice of $\epsilon$ is different}
 By considering two different cases ($\Bex=\Omega(\PTP)$ and $\Bex=O(\PTP)$) and choosing $\epsilon$ accordingly,
 %\bo{it would be easier for the reader to see if you can explain how to choose $\epsilon$ and give the choice of $\epsilon$. otherwise, it would also be harder to follow the insights discussed below}\duo{the following paragraphs explain how $\epsilon$ is chosen and leads to different lower bounds. Do we need to state here again?}
 we show that one of $\mathbf{B.2}$ and $\mathbf{B.3}$ dominates the other. Then, we can obtain the desired lower bound by combining these two cases. This completes the proof sketch.
%\duo{added one remark}
\end{proof}
\begin{remark}\label{remark:purdue}
    While we use the same instance construction method in \citet{shi2022power}, the problem they study is very different from ours. In particular, their learning protocol and the definition of switching costs are different, and they do not consider an observation budget as we do. We present a detailed discussion about the key difference in Section~\ref{sec:related_work}.%\xingyu{it is not easy for understanding. You can just say that their problem is different in terms of problem setup and regret definition. Then, refer to Appendix for further discussion. No need to give details here}\duo{done}
\end{remark}

\subsection{Insights from Lower Bound Analysis}\label{sec:lower_bound_insight}
Next, we give some useful observations and important insights that can be obtained from the above proof sketch, in particular, from \eqnref{extra_lower_bound_before_minimize}, which provides a \emph{unified} view of the lower bound in online learning with bandit feedback and flexible extra observations within a budget.

As a warm-up, we begin with the standard bandit case (i.e., $\Bex = 0$), which has been extensively studied~\cite{dekel2014bandits}. Recall that under the current instance construction, bandit feedback provides useful information \emph{only} when the learner switches actions. From \eqnref{extra_lower_bound_before_minimize}, one can observe that there is a tradeoff between exploration and switching costs: on the one hand, in order to better explore and enjoy a lower regret, the learner has to switch frequently (i.e., a larger $S$) so as to gain more information (i.e., a larger $\mathbf{A.2}$); on the other hand, however, since the learner has to pay one unit of switching cost for each switch (contributing to $\mathbf{A.3}$), she should not switch too often. 
%For example, if the learner switches $T$ times, the regret would already be $\Omega(T)$ purely due to the switching costs. 
To strike the balance between the two, the best the learner can do is to switch $S^* := \Theta(\PTP)$ times; otherwise, the regret can only be worse because $S^*$ is the minimizer of the lower bound in \eqnref{extra_lower_bound_before_minimize}. Finally, choosing $\epsilon$ to be $\wtTheta(K^{1/3}T^{-1/3})$ in \eqnref{lower_bound_after_minimize} yields the $\wtOmega(\PTP)$ bound for the bandit case.
\begin{remark}\label{remark:key_observation}
    The above discussion indicates that with switching costs, the worst-case hard instance restrains the learner from obtaining distinguishability from  more than $\Theta(\PTP)$ rounds (i.e., rounds associated with action switches) rather than $T$ rounds as in the standard bandit learning setting (without switching costs). This is also the key reason why the minimax regret is worse in bandit learning with switching costs. 
\end{remark}
\vspace{-3mm}
Next, we consider the first case: $\Bex=O(\PTP)$. In this case, one might hope to obtain a smaller regret (compared to the bandit case) with the help of additional feedback.
%However, after going through how we obtain the regret lower bound of $\wtOmega(\PTP)$ for the bandit case, it might be easier to see why $\wtO(\PTP)$ extra observations are insufficient to improve the minimax regret: intuitively, although the learner can now make some extra observations that help obtain additional information, the gain is too small to improve the regret order-wise.
However, we will show that unfortunately, the gain from those additional observations is negligible for improving the regret order-wise, and hence, the previous $\wtOmega(\PTP)$ bound remains. To see this, let $\epsilon$ take the same value as in the bandit case (i.e., $\epsilon=\wtTheta(K^{1/3}T^{-1/3})$)  in Eq. (\ref{eqn:lower_bound_after_minimize});
although $\mathbf{B.3}$ now becomes positive instead of zero (as in the bandit case), it is still dominated by $\mathbf{B.2}$, which results in the same $\wtOmega(\PTP)$ bound as in the bandit case.

%\bo{this part is unclear and needs to be rewritten}\duo{rewritten below} Finally, we consider the second case: $\Bex = \wtOmega(\PTP)$. In this case, extra observations are sufficient in the sense that the regret  help improve the regret to $\wtOmega(T\sqrt{K/\Bex})$. Consequently, by a different choice of $\epsilon = \wtTheta(\sqrt{K/\Bex})$, now $\mathbf{B.3}$ is dominating $\mathbf{B.2}$ and we arrive at our desired result. That is to say, now the improved bound comes from better exploration (which corresponds to $\mathbf{A.2}$) due to a sufficient amount of extra observations (which however will not incur extra switching costs) so that the learner now does not have to necessarily switch $\wtTheta(T^{2/3})$ times in order to explore well.

%\duo{rewrite the $\uparrow$ paragraph}
We now turn to the second case: $\Bex = \Omega(\PTP)$. In contrast to the previous case, due to a relatively large budget, the distinguishability provided by those extra observations (which do not contribute to switching costs) is no longer negligible. This leads to a smaller regret. In particular, by choosing $\epsilon = \wtTheta(\sqrt{K/\Bex})$, we have $\mathbf{B.3}$ dominate $\mathbf{B.2}$ and obtain the desired lower bound. In other words, one can reduce the regret through free exploration enabled by such extra observations without incurring switching costs.
%\bo{use the integrated tool Grammarly in overleaf to help you spot and fix such minor editorial issues (typos, misspellings, grammatical mistakes, etc.); make sure to think about the suggestions provided by the tool though}\duo{sure, typically I'll try to figure it out whenever I see an error suggested.}
%\xingyu{you might need to write the previous paragraph (when extra obs is small) like this paragraph. That is, first give the intuition and then give the math.} \duo{done}
%\textbf{Connections to upper bound and algorithm design.} 

\subsection{Fundamental Questions about Algorithm Design}
\label{sec:questions}
% \duo{i'm thinking of use another word other than ``insights'' to avoid using it twice}}\xingyu{or simply algorithm design?}

The above insights we gain from the lower bound analysis can also shed light on the algorithm design. In fact, these motivate us to ask several fundamental questions, not only about how to achieve optimal regret but also about the role of feedback in online learning with switching costs, in terms of both the amount and type of feedback.   

% In the following, we also provide useful insights into the design of learning algorithms that can not only match this lower bound, but also enjoy improved sample efficiency if possible. 
% So far, we have derived a good understanding of the lower bound, yet it is still unclear how to design learning algorithms that can achieve this lower bound.

On the one hand, it is straightforward to achieve a matching upper bound when $\Bex=O(\PTP)$. Specifically, 
%based on our insight above,\duo{why is it related to the insight above? batched EXP3 is an existing algorithm and we can just use it.} 
one can simply ignore all the extra observations and use bandit feedback only, e.g., batched EXP3~\cite{Dekel2012OnlineBL}, which enjoys a $\wtTheta(\PTP)$ regret.
%\bo{use $\wtTheta$?} \duo{I thought whenever we mention the performance guarantee of an alg. we should use O instead of Theta. Moreover, on some easier instances the regret can be of lower order, so I don't see why use $\Theta$}\bo{we are talking about instance independent bound here, right? Since it achieves the lower bound, it should be fine to use $\wtTheta$}\duo{my point is that, it's not achieving this regret on any instance, but can be smaller. when we say an alg's regret, it's not minimax regret.}
Although the bounds match,
only $\Theta(T^{2/3})$ of the bandit feedback from the $T$ rounds contribute to distinguishability due to the tradeoff introduced by switching costs (see Remark~\ref{remark:key_observation}).
%, while the rest just becomes redundant as they do not provide any useful information (see Remark~\ref{remark:key_observation}).
Given this observation, it is natural to ask: 
\emph{\textbf{(Q1)} Can one still achieve the same regret of $\wtTheta(\PTP)$ while using bandit feedback from $\Theta(\PTP)$ rounds only? 
%\bo{$\Theta$ or $\wtTheta$?}\duo{motivated by lower bound, it should be $\wtTheta$} 
%\textbf{(Q2)} 
Moreover, how would regret scale with the amount of available feedback if the (bandit) feedback is even more limited (e.g., $O(\PTP)$)?}
%\xingyu{bandit feedback?}\duo{yes. fixed}

% On the other hand, it remains largely unknown how to match the lower bound of $\wtOmega(T\sqrt{K/\Bex})$ 
% when $\Bex = \wtOmega(\PTP)$. Note that in the derivation of the lower bound, we \emph{optimistically} treat \emph{all} $\Bex$ extra observations as contributing to useful information toward distinguishability (see term~$\mathbf{A.2}$ in \eqnref{extra_lower_bound_before_minimize}). To achieve this, however, one needs to answer an important question: \emph{\textbf{(Q3)} How to carefully design a learning algorithm that properly uses these extra observations to indeed gain sufficient useful information toward distinguishability and match the lower bound?}

%\duo{I guess I deleted this paragraph by mistake before...}\bo{should be able to find it back in history}\duo{now it's here, found it in my backup downloaded}\bo{in history, you can add a label for some important versions}\duo{Isee. not aware of this function. it's exactly the paragraph commented out above. so not sure if I deleted it by mistake or someone thought it's not needed.}\bo{I think we need it. I added the above (commented out) paragraph just now by copying and pasting from a version in the history.}
On the other hand, it remains largely unknown how to match the $\wtOmega(T\sqrt{K/\Bex})$ bound
when $\Bex = \Omega(\PTP)$. Note that in the derivation of the lower bound, we \emph{optimistically} view that \emph{all} $\Bex$ extra observations contribute to useful information toward distinguishability 
%\duo{treat sth. as doing...? if no grammatical errors please delete my comment} 
(see term~$\mathbf{A.2}$ in \eqnref{extra_lower_bound_before_minimize}). To achieve this, however, one needs to answer an important question: \emph{\textbf{(Q2)} How to carefully design a learning algorithm that can properly use these extra observations to indeed gain sufficient useful information toward distinguishability and match the lower bound?
Moreover, since $\Bex$ now dominates $S^*$ (order-wise), %a further interesting question is: 
%\emph{\textbf{(Q4)} Can 
can one still match the lower bound of $\wtOmega(T\sqrt{K/\Bex})$ using $\Bex$ extra observations only (i.e., not using any bandit feedback)?}

To address these fundamental questions, it turns out that it would be more instructive to consider a general setting where the learner has a budget for total observations (see Section~\ref{sec:total}) rather than extra observations. We will show that the results obtained for this general setting will naturally answer the aforementioned questions. In particular, we show that there exist learning algorithms that can match the lower bound (up to poly-logarithmic factors), hence concluding the minimax regret stated in Theorem \ref{thm:bandit_extra_minimax}.

% Motivated by the above questions, we come to realize that the answers to them could be answered and captured as special cases in a more general setting, where the learner makes feedback within \emph{a total budget}. In particular, the answer to $\mathbf{Q1}$ could be viewed as the optimal regret when the learner exploits her total budget (which is no larger than $T$) in a certain way (namely, bandit feedback). And that of $\mathbf{Q2}$ is just a special case when the total budget is $\wtOmega(\PTP)$.

%% file: sec5-xingyu.tex
\section{Online Learning with Switching Costs under Total Observation Budget}\label{sec:total}

%\xingyu{Hi all, all comments in this section look good to me. Feel free to make your changes directly:)}

In this section, we consider a more general setting of online learning with switching costs under a total observation budget. Specifically, at each round, the learner can freely choose to observe the loss of up to $K$ actions (which may not necessarily include the action played), as long as the total number of observations over $T$ rounds does not exceed the budget $B$, which is an integer in $[K, KT]$. Without loss of generality, we assume $B \geq K$.
%\footnote{Without loss of generality, we assume $B \geq K$.}
% \footnote{Without loss of generality, we assume $B \geq K$. Otherwise, the learner cannot even observe the loss of some action at least once, thus resulting in linear regret in the worst case.} 
We aim to understand the role of feedback in this general setting
by studying the following fundamental question: 
\emph{\textbf{(Q3)} How does the minimax regret scale with the amount of available feedback in general? 
What is the impact of different types of feedback (bandit, full-information, etc.)?}
% by studying two fundamental questions: 
% \emph{\textbf{(Q3)} How does regret scale with the amount of available feedback in general? 
% \textbf{(Q4)} What is the impact of different types of feedback (bandit, full-information, etc.)?}

%, which not only enjoys its own interest, but helps to answer our raised important questions in the last section. 

% by studying two fundamental questions: 
% \emph{\textbf{(Q5)} How does regret scale with the amount of available feedback in general? \textbf{(Q6)} What is the impact of different types of feedback (bandit, full-information, etc.)?}
% \bo{not sure if this will be too many}\duo{do we need to call back? if no then maybe no need to label them}\bo{I think we should call back after presenting the theorems/propositions.}

% To this end, we first present the minimax regret in this setting, which captures the optimal scaling with the amount of feedback $B$, thus answering question \textbf{(Q5)}. Then, we propose a generic algorithmic framework that enables us to design flexible algorithms to achieve optimal regret. Finally, we discuss the role of feedback type in the design of algorithms, which offers answers to question \textbf{(Q6)} as well as those raised at the end of Section~\ref{sec:extra}. \bo{may consider deleting this paragraph if need space; in this case, we need to refer back to Q5 and Q6 later}
% \bo{I now tend to comment out this paragraph}

To proceed, we need some additional notations for this section. Let $\mathcal{O}_{t} \subseteq [K]$ be the observation set, i.e., the set of actions whose loss the learner chooses to observe at round $t \in [T]$, and let $N_{\mathrm{ob}}$ be the total number of observations, i.e., $N_{\mathrm{ob}} :=  \sum_{t=1}^T |\mathcal{O}_{t}|$.
Naturally, we have $N_{\mathrm{ob}} \le B \le KT$. For example, bandit feedback is a special case with $\mathcal{O}_{t} = \{X_t\}, \forall t\in[T]$ and $N_{\mathrm{ob}} = B = T$; full-information feedback is another special case with $\mathcal{O}_{t} = [K], \forall t\in[T]$ and $N_{\mathrm{ob}} = B = KT$.
By slightly abusing the notation in this section, we also use $R^*_T$ to denote the minimax regret over the set of all learning algorithms that satisfy the learning protocol specified in Section~\ref{sec:problem setup} and do not exceed the total observation budget $B$.

%Now we sightly abuse the notation and let $\Pi$ denotes the set of all the learning algorithms making $B$ observations in total.
%\bo{Duo, you need to be very careful when addressing comments. There are a lot of editorial issues in your modified test; see this paragraph for example. We may not have enough time.}

\subsection{Minimax Regret}
We first present the main result of this section and fully characterize the minimax regret for this general setting.
%, i.e., online learning with switching costs under a \emph{total} observation budget.  

% First, we present a general lower bound under any budget $B \in [K: KT]$ \bo{I do not quite like this notation, which is not straightforward. As long as you make it clear that $B$ is a positive integer, it is fine to use $B \in [K,KT]$}\xingyu{I agree}. This result is stated in \thmref{yasin_lower_bound} below.

% \begin{theorem}
% The minimax regret of online learning with switching costs under a total observation budget $B \in [K,KT]$ is the following: $R_T^*=\widetilde{\Theta}(T\sqrt{K/B})$.   \label{thm:total_minimax}
% %$\wtOmega( \frac{T\sqrt{K}}{\sqrt{B}})$.
% \end{theorem}

\begin{theorem}\label{thm:total_minimax}
In the setting of online learning with switching costs under a total observation budget $B \in [K,KT]$, the minimax regret is given by $R_T^*=\wtTheta(T\sqrt{K/B})$.   
\end{theorem}

\begin{remark}
    This result answers the first part of question \textbf{(Q3)}: the minimax regret has a universal $\Theta(1/\sqrt{B})$ scaling across the full range of total budget $B$ (see Fig.~\ref{fig:main_results} (b)), compared to the phase transition in~\cref{sec:extra} (see Fig.~\ref{fig:main_results} (a)).
\end{remark}
 To establish this result, we need to obtain both a lower bound and a matching upper bound. For the lower bound, it turns out that it suffices to use an existing lower bound, which was originally
derived for standard online learning \emph{without} switching costs. We restate this lower bound in Lemma~\ref{lem:total_lower}.

\begin{lemma}\citep[Theorem~2]{seldin2014prediction}
\label{lem:total_lower}
In the setting of online learning (without switching costs) under a total observation budget $B \in [K,KT]$, the minimax regret is lower bounded by $R_T^*={\Omega}(T\sqrt{K/B})$.
\end{lemma}

Naturally, this serves as a valid lower bound for the setting with switching costs we consider. In fact, we will show that this lower bound is tight (up to poly-logarithmic factors), which in turn offers the following important message. 

\begin{remark}
If the learner can freely make observations over $T$ rounds within the budget, introducing switching costs \emph{does not increase} the intrinsic difficulty of the online learning problem in terms of the minimax regret.
\end{remark}

Now, it only remains to show that there exist algorithms that can achieve a matching upper bound (up to poly-logarithmic factors), which will be the main focus of the next subsection.

\subsection{Learning Algorithms and Upper Bounds}\label{sec:full and flex alg}

In this subsection, we show that there indeed exist algorithms that can achieve the lower bound in Lemma~\ref{lem:total_lower}, which further implies the tight bound in Theorem~\ref{thm:total_minimax}. Instead of focusing on one particular algorithm, we first propose a generic algorithmic framework, which not only enables us to design various optimal learning algorithms in a unified way but also facilitates a fundamental understanding of the problem by distilling its key components.  

% To this end, we first step back and propose a generic algorithmic framework, which not only allows us to  
% \xingyu{continue to work from here...}
%\bo{Rephrase this paragraph as a remark?}
% \begin{remark}
%  Note that the above lower bound was originally derived for standard online learning \emph{without} switching costs; it has been shown to be tight (up to a multiplicative $\sqrt{\ln K}$ factor) in the standard setting~\cite{seldin2014prediction}. Clearly, it also serves as a valid lower bound for online learning \emph{with} switching costs we consider. Yet, this lower bound may be loose as introducing switching costs can potentially result in an increased regret rate. A little surprisingly, however, in the sequel we will show that this lower bound is also tight (still up to a $\sqrt{\ln K}$ factor) with switching costs.
% \end{remark}

%\bo{As I suggested to you, one option is to state the upper bound result in Theorem 3 using $R^*_T$ as well; do not mention any algorithm in Theorem 3. Now, I think we can state Theorem 3 here. Then, you can have a separate subsection focused on the algorithms.}

% \xingyu{rewrite all the parts below until Theorem 3. Please check.}

% To enjoy matching upper bound and characterize the minimax regret, we propose an algorithmic framework, of which we will introduce several instantiations later in this section. 

Our generic framework builds upon the classic \emph{Online Mirror Descent (OMD)} framework with negative entropy regularizer (also called the \emph{Hedge} algorithm)~\citep{Littlestone1989TheWM} and incorporates the following three key components to tackle both switching costs and observation budget in a synergistic manner. 

% The complete pseudo-code is given by Algorithm~\ref{alg:framework}.
% Overall, our algorithm design takes Online Mirror Descent (OMD) with negative entropy regularizer (also named \emph{Hedge} in the literature) as a sub-routine and runs it over batches, combined with some other techniques (e.g., Shrinking Dartboard~\citep{geulen2010regret}). Before we give a detailed description of our algorithm, let us first highlight its three key components. 

\textbf{Batching Technique.} The batching technique was originally proposed for addressing adaptive adversaries~\citep{Dekel2012OnlineBL}, but naturally provides low switching guarantees. We divide $T$ rounds into batches and judiciously distribute the available observations across batches. That is, instead of consuming observations at every round as in standard online learning (which could even be infeasible when observation budget $B$ is relatively small), we use observations only at a \emph{single} round randomly sampled from each batch. One key step to obtain the desired regret guarantee is to feed the (unbiased estimate of) batch-average loss to the learning algorithm at the end of each batch. While this technique is borrowed from \citet{shi2022power}, the problem setup we consider is very different (see Section~\ref{sec:related_work}).
%\bo{be more specific, e.g., they do not consider budget. I will add something here.}\duo{we have remark \ref{remark:purdue} earlier in sec 4. should we mention all the differences there? I prefer to remove this sentence}

\textbf{Shrinking Dartboard (SD).} SD is a calibrated technique for controlling the number of action switches in online learning under a lazy version of Hedge. That is, with a carefully crafted probability distribution, the action tends to remain unchanged across two consecutive rounds~\cite{geulen2010regret}
%\duo{$\leftarrow$I don't understand this sentence. what is ``using''?} 
while preserving the same marginal distribution as in Hedge. In our algorithmic framework, we generalize this idea to the batching case with general feedback: the same action can be played across two consecutive batches (instead of across rounds), and it is no longer required to use only full-information feedback as in~\citet{geulen2010regret}.

\textbf{Feedback Type. }
%\bo{how about simply ``Feedback Type'' or ``Feedback Selection''?}\duo{I think we can choose between ``attainment'' and ``type''. ``selection'' is weird to me}
Recall that the learner is allowed to freely request feedback within the total budget. Hence, our last component lies in the feedback type. That is, the learner has the flexibility to choose the observation set $\mathcal{O}_{u_b}$ (not limited to bandit or full-information feedback only). In order to achieve a matching upper bound, however, the choice of the observation set (i.e., the type of feedback) is crucial in some cases. 
%and could even be potentially restricted \bo{what does it mean?}\duo{less choices, i.e., in some cases, one cannot exploit the budget by bandit feedback (otherwise suffer sub-optimal regret)}\bo{I'd prefer to simply not say it here.}. 
We will elaborate on this in Section~\ref{subsec:feedback_type}. 

% Therefore, we leave the choice of set $\mathcal{O}_{u_b}$ as a variable component. In fact, 

Putting these three components together, we arrive at our unified algorithmic framework, which is presented in Algorithm~\ref{alg:framework}. Given the input $T$, $K$, and $B$ of the problem, we need to determine the following input of the algorithm: the number of batches $N$, batch size $\tau$, learning rate $\eta$, and indicator $I_{\mathrm{SD}}$ ({Line~\ref{line:alg1_input}}), along with the initialization of some variables ({Line~\ref{line:alg1_initial}}).
%\bo{these are all input, but I think we should separate $B$, $K$ and $T$ from others in the algorithm, which will be chosen according to $B$, $K$ and $T$}\duo{put them after initialization?}\bo{I prefer to keep them in the input, but my point is that we should separate them and make it clear that they are determined based on $B$, $K$ and $T$}
Throughout the game, we maintain a positive weight $W_b[k]$ for each action $k\in[K]$ in each batch $b\in[N]$. Both the weights and the action for each batch may be updated only between two consecutive batches. Hence, in each batch $b$, we keep playing the chosen action $A_b$ until the end of the batch ({Line~\ref{line:alg1_take_action}}); we sample a round $u_b$ uniformly at random from the current batch ({Line~\ref{line:alg1_choose_time_slot}}) and choose an observation set $\mathcal{O}_{u_b}$ in a certain way (to be specified later) such that the loss of each action in $\mathcal{O}_{u_b}$ will be observed at round $u_b$ ({Line~\ref{line:alg1_observe}}). We then construct an unbiased estimate ({Line~\ref{line:alg1_loss_estimate}}), denoted by $\widehat{\ell}_b=(\widehat{\ell}_b[1], \dots, \widehat{\ell}_b[K])$, of the batch-average loss $\sum_{t=(b-1)\tau+1}^{b\tau}\ell_t/\tau$ (which depends on the choice of $\mathcal{O}_{u_b}$ and will be specified later) and then update the weight and sampling probability of each action accordingly: $W_{b+1}[k] = W_{b}[k] \cdot \exp(-\eta \cdot \widehat{\ell}_b[k])$ and $w_{b+1}[k]:=W_{b+1}[k]/\sum_{i=1}^K W_{b+1}[i]$ ({Line~\ref{line:alg1_OMD_update}}).
%\bo{then it is inconsistent - here, you indeed determine the action for next batch $b+1$}\duo{anything wrong now?I'm talking about how to determine $A_b$ and get rid of $p_b$}
% \duo{Now we state how next decision $A_{b+1}$ is determined when $I_{\mathrm{SD}}=0$ and $1$ respectively (Line~9).} 
Finally, we determine action $A_{b+1}$ for the next batch ({Line~\ref{line:next_decision}}). Specifically, if the SD indicator $I_{\mathrm{SD}}=0$, probability $I_{\mathrm{SD}} \cdot \exp(-\eta \cdot \widehat{\ell}_b)$ is always zero, and hence, action $A_{b+1}$ is sampled using fresh randomness with probability proportional to action weights as normally done in Hedge: sample $A_{b+1}$ following distribution $w_{b+1} = (w_{b+1}[1], \dots, w_{b+1}[K])$. If the SD indicator $I_{\mathrm{SD}}=1$, with probability $\exp(-\eta \cdot \widehat{\ell}_b)$, we keep the current action for the next batch (i.e., $A_{b+1}= A_b$); otherwise, we sample a new action $A_{b+1}$ following distribution $w_{b+1}$.

\begin{algorithm}[t!]
    \caption{Batched Online Mirror Descent with 
    %Negative Entropy Regularizer \bo{I think no need to specify the regularizer here}\duo{I agree} and 
    (Optional) Shrinking Dartboard} %(BOMD-SD)\bo{Do we ever use the acronym?}\duo{I don't think so}\xingyu{then, not necessary}}
    \label{alg:framework}
    \begin{algorithmic}[1]
        \STATE \textbf{Input:} length of time horizon $T$, number of actions $K$, and observation budget $B$; determine the following based on $T$, $K$, and $B$: number of batches $N$, batch size $\tau=T/N$, learning rate $\eta$, and SD indicator $I_{\mathrm{SD}}$\alglinelabel{line:alg1_input}
        \STATE \textbf{Initialization:} action weight $W_1[k] = 1$ and action sampling distribution $w_1[k] = 1/K, \forall k \in [K]$; $\mathcal{O}_t =\emptyset, \forall t\in [T]$; choose $A_1 \in [K]$ uniformly at random \alglinelabel{line:alg1_initial}
        
        \FOR {batch $b = 1: N$}
            \STATE Play action $A_b$ throughout the current batch $b$, i.e., $X_t = A_b, \forall t = (b-1)\tau+1 ,\dots, b\tau$ \alglinelabel{line:alg1_take_action}
            \STATE Sample a round index $u_b$ uniformly at random from integers in $[(b-1)\tau+1, b\tau]$  \alglinelabel{line:alg1_choose_time_slot}
            \STATE Choose an observation set $\mathcal{O}_{u_b}\subseteq[K]$ (to be specified later) and observe the loss of each action in $\mathcal{O}_{u_b}$ at round $u_b$: $\{\ell_{u_b}[i]: i\in \mathcal{O}_{u_b}\}
            $\alglinelabel{line:alg1_observe}
            \STATE Construct unbiased estimate $\widehat{\ell}_b$ (to be specified later) of the batch-average loss $\sum_{t=(b-1)\tau+1}^{b\tau}\ell_t/\tau$\alglinelabel{line:alg1_loss_estimate}
            \STATE Run OMD update: update the weight of each action: $W_{b+1}[k] = W_{b}[k] \cdot \exp(-\eta \cdot \widehat{\ell}_b[k])$ and the sampling probability: $w_{b+1}[k] = \frac{W_{b+1}[k]}{\sum_{i=1}^{K}W_{b+1}[i]}, \forall k\in[K]$\alglinelabel{line:alg1_OMD_update}
            \STATE With probability $I_{\mathrm{SD}} \cdot \exp(-\eta \cdot \widehat{\ell}_b)$, keep action $A_{b+1} = A_b$; otherwise, sample action $A_{b+1}\sim w_{b+1}$\alglinelabel{line:next_decision}
        \ENDFOR
    \end{algorithmic}
\end{algorithm}
With Algorithm~\ref{alg:framework} in hand, we are ready to introduce several specific instantiations and study their regret guarantees. In particular, for each instantiation we will specify the choice of the following parameters: number of batches $N$, batch size $\tau$, learning rate $\eta$, SD indicator $I_{\mathrm{SD}}$, and observation set $\mathcal{O}_{u_b}$. In the following, we first demonstrate one simple instantiation that uses full-information feedback only. Then, we show how to generalize this instantiation using more flexible feedback (i.e., not limited to full information only) while achieving the same performance guarantee.

\textbf{Instantiation via Full-information Feedback.} In this instantiation of Algorithm~\ref{alg:framework}, we receive full-information feedback at a randomly selected round $u_b$ in each batch $b$ (i.e., $\mathcal{O}_{u_b}
 = [K]$ and $\widehat{\ell}_b = \ell_{u_b}$)
 %\bo{is this consistent with line 7? same comment for Proposition~1}\duo{yes, in this case, we observe the losses of all actions at $u_b$, so we do have $\ell_{u_b}$ and we do use it as our estimate}
 and SD is turned on (i.e., $I_{\mathrm{SD}} = 1$). At a high level, this can be viewed as a batched generalization of the original SD algorithm \cite{geulen2010regret} with $N=B/K$ batches
 %\footnote{For ease of exposition, we assume that $N$ and $\tau$ are integers.}
 %\footnote{For ease of exposition, we assume that $N$ and $\tau$ are integers. In Appendix~\ref{app:integer}, we discuss how to deal with the general case.}
 % \footnote{Since we are interested in the asymptotic regret bounds (i.e., when both $K$ and $T$ are sufficiently large), we ignore all the \emph{integer-related} issues. However, in Appendix \ref{app:integer} we will show how to deal with this while still enjoying the same (asymptotic) regret bounds.}
 %\bo{you need to add ceiling function or explicitly mention that we assume it is an integer for ease of presentation}\duo{I made a footnote saying ``Since we are interested in the asymptotic regret bounds (i.e., both $K$ and $T$ are sufficiently large), we will ignore all the \emph{integrality-related} issues throughout this section. However in Appendix \ref{app:integer} we show how to deal with them while still enjoying the same (asymptotic) regret bound.'' which was deleted}
(since we have $K$ observations in each batch), and hence, the corresponding batch size is $\tau=T/N=KT/B$.  
For ease of exposition, we assume that $N$ and $\tau$ are integers.
%In particular, this instantiation (denoted by $\pi_1$) can be specified by: 
Specifically, we have
$N=B/K$, $\tau = KT/B$, $\eta = \sqrt{\frac{2\ln K}{3B}}$, $I_{\mathrm{SD}}=1$, $\mathcal{O}_{u_b}=[K]$, and $\widehat{\ell}_b =\ell_{u_b}$. We use $\pi_{\mathrm{full}}$ to denote this instantiation and present its regret upper bound in Proposition~\ref{prop:uniform_alg_upper_bound}. The proof is provided in Appendix~\ref{app:uniform_alg_upper_bound_proof}.
%\duo{For the complete proof, please see Appendix \ref{app:uniform_alg_upper_bound_proof}.} 

\begin{proposition}\label{prop:uniform_alg_upper_bound}
The worst-case regret under algorithm $\pi_{\mathrm{full}}$ is upper bounded by $R_T^{\pi_{\mathrm{full}}} = O(T\sqrt{K\ln K/B})$.
%\bo{$R_T$ is not defined. Do you want to give a name to this instantiation so that you can use the notation in Eq.~(2)?}\duo{how about now}
\end{proposition}

 \begin{remark}
     This result immediately implies an upper bound of the minimax regret: $R^*_T = O(T\sqrt{K\ln K/B})$, which, along with the lower bound in Lemma~\ref{lem:total_lower}, further implies the tight bound in Theorem~\ref{thm:total_minimax}. Note that there is an additional $\sqrt{\ln K}$ factor in the upper bound. This shares the same pattern as in the setting even without switching costs (see \citet[Theorem~1]{seldin2014prediction}), where the achieved upper bound also has an additional $\sqrt{\ln K}$ factor.
     %\xingyu{shall we mention other regularizer?}\bo{I think it is good to mention that.}\duo{please see remark~\ref{remark:improved_bandit}}
 \end{remark}

%\bo{rewrote the following remark; please check} \xingyu{good to me}
  \begin{remark}\label{remark:call_back_extra_minimax}
     For the previous setting considered in Section~\ref{sec:extra}, the above result also implies an upper bound of the minimax regret: $\wtO(T\sqrt{K/\Bex})$, when $\Bex=\Omega(\PTP)$, by simply ignoring all bandit feedback (i.e., $B=\Bex$).
     On the other hand, as discussed in Section~\ref{sec:questions}, when $\Bex=O(\PTP)$, one can simply ignore extra observations and use pure bandit feedback only (e.g., batched EXP3~\cite{Dekel2012OnlineBL}) to achieve a $\wtO(\PTP)$ regret.
     Combining these results, along with the lower bound in Proposition~\ref{prop:bandit_extra_lower_bound}, implies the tight bound in Theorem~\ref{thm:bandit_extra_minimax}.
     Moreover, this also answers question $\textbf{(Q2)}$ raised in Section~\ref{sec:extra}.
 \end{remark}

The result of our first instantiation shows that the optimal regret can indeed be achieved (up to a $\sqrt{\ln K}$ factor) when full-information feedback is employed. However, we can also show that the use of full-information feedback is not essential. In fact,
%\bo{this sentence is not quite clear}
it suffices to have an observation set chosen uniformly at random from all subsets of $[K]$ with the same cardinality, which leads to a more flexible instantiation of Algorithm~\ref{alg:framework} presented below. 

\textbf{Instantiation via Flexible Feedback.} In this instantiation, instead of having $|\mathcal{O}_{u_b}|
 = K$ as under full-information feedback, we allow $|\mathcal{O}_{u_b}|
 = M \le K$. The key to this flexibility is a careful construction of an unbiased estimate of the batch-average loss (i.e., $\widehat{\ell}_b$).
 %, which is explicitly given by the statement of the following result. 
 Specifically, let $M$ be any integer that satisfies $M \in [K]$ if $B<T$ and $M \in [\lceil B/T \rceil, K]$ if $B \ge T$.\footnote{To fully use the budget, $M$ cannot be too small when $B \geq T$.}
 % \footnote{When $B\geq T$, the value of $M$ cannot be too small. Otherwise, the available observations will not be fully used, which results in suboptimal regret.}
 Then, we have
 % this instantiation (denoted by $\pi_2$) can be specified by: 
 $N=B/M$, $\tau = T/N = MT/B$, $\eta = M\sqrt{\frac{2\ln K}{3KB}}$, $I_{\mathrm{SD}}=1$, $\mathcal{O}_{u_b}$ is chosen uniformly at random from $\{U\in 2^{[K]}: |U|=M\}$, and $\widehat{\ell}_b[k] = \bI\{k\in \mathcal{O}_{u_b}\} \cdot \frac{\ell_{u_b}[k]}{M/K}$ for all $k\in[K]$.
 We use $\pi_{\mathrm{flex}}$ to denote this instantiation and present its regret upper bound in Proposition~\ref{prop:flex upper bound}. The proof is provided in Appendix~\ref{app:B<T_upper_bound_proof}.
 %\duo{We leave the complete proof to Appendix \ref{app:B<T_upper_bound_proof}}
 %, where $M$ is some integer specified below in Proposition \ref{prop:flex upper bound}.

% The short answer is yes, although it in fact may depend on the value of $B$ and differ a little bit. Let us use $M\in[K]$ to denote the number of observations we make each time (i.e., $M=|\mathcal{O}_{u_b}|$). We formally state the flexibility regarding $M$ in the following proposition. Since there is nothing special in the algorithm design and regret analysis on top of Proposition \ref{prop:uniform_alg_upper_bound}, we leave the detailed discussions and proof to Appendix \ref{app:B<T_upper_bound_proof}.

\begin{proposition}\label{prop:flex upper bound}
% Let $M$ be any integer that satisfies
% \begin{equation*}
%         M \in \left\{
%         \begin{aligned}
%         &[K] , & \text{if} ~B<T, \\
%         &[\lceil B/T \rceil, K] , & \text{if}~ B\geq T.
%         \end{aligned}
%         \right.
%     \end{equation*}
%\bo{Why dose it have to be so complicated? Why not simply $M \in [K]$? This big bracket expression would look awkard.}
The worst-case regret under algorithm $\pi_{\mathrm{flex}}$ is upper bounded by $R_T^{\pi_{\mathrm{flex}}} = O(T\sqrt{K\ln K/B})$.
%\bo{I feel it would be better to describe the instantiation before the proposition and give a name. Then the statement of the proposition can be much simplified. Basically, Algorithm XXX achieves $\dots$}\duo{how about now}
\end{proposition}

An astute reader may already notice that in the above flexible instantiation, while the number of observations can be one (i.e., $|\mathcal{O}_{u_b}|
 = 1$), it is not the same as standard bandit feedback. This is because here, $\mathcal{O}_{u_b}$ needs to be chosen uniformly at random rather than simply being the action played in that batch (i.e., $\mathcal{O}_{u_b} = \{A_b\}$) as in the standard bandit setting (with a batch size of one). Motivated by this subtle difference, we will devote the next subsection to studying the impact of feedback type.
 %the effect of bandit feedback.
%\bo{I am thinking more of the limitation of bandit feedback}\duo{by potential we mean it's not always useless.}

%\bo{I modified the following subsection. Please check.}

\subsection{Impact of Feedback Type}\label{subsec:feedback_type}

%In this subsection, we consider learning algorithms that use bandit feedback only. Hence, we have $B \le T$. 
%We show that when the budget $B$ is limited (i.e., $B = O(\PTP)$), using bandit feedback can still achieve the $\Theta(1/\sqrt{B})$ scaling, but when the budget $B$ is large (i.e., $B=\Omega(\PTP)$), bandit feedback is \emph{no longer sufficient} to achieve optimal regret. Clearly, this illustrates the crucial impact of the type of feedback, especially when the budget $B$ is large.

In this subsection, we study the impact of feedback type by presenting another instantiation of Algorithm~\ref{alg:framework} via pure bandit feedback only. In this case, we naturally have $B \le T$.

\textbf{Instantiation via Bandit Feedback.} This instantiation is a generalized version of batched EXP3 \citep{Dekel2012OnlineBL} with \emph{flexible batch size}.
Specifically, we have $N=B$,  $\tau = T/B$, $\eta = \sqrt{\frac{2\ln K}{BK}}$, $I_{\mathrm{SD}} = 0$, $\mathcal{O}_{u_b}=\{A_b\}$, and $\widehat{\ell}_b[k] = \bI\{k\in \mathcal{O}_{u_b}\} \cdot \frac{\ell_{u_b}[k]}{w_b[k]} $ for all $k\in [K]$. 
 We use $\pi_{\mathrm{b}}$ to denote this instantiation. When $B=O(\PTP)$,
 we obtain a regret upper bound for $\pi_{\mathrm{b}}$ and state it in Proposition~\ref{prop:bandit_upper_bound}. The proof is provided in Appendix~\ref{app:bandit_upper_bound_proof}.

%With this, its regret guarantee is given by the following result. 

% Although \citet{Dekel2012OnlineBL} and \citet{dekel2014bandits} altogether imply the $\wtTheta(\PTP)$ optimal regret when $B=T$ under bandit feedback, a tight characterization of the minimax regret for general $B\leq T$ is still unclear.

% To close this gap, we show another instantiation from our Algorithm \ref{alg:framework}, together with its performance guarantee in the Proposition below.

\begin{proposition}\label{prop:bandit_upper_bound}
When $B=O(\PTP)$, the worst-case regret under algorithm $\pi_{\mathrm{b}}$ is upper bounded by $R_T^{\pi_{\mathrm{b}}} = O(T\sqrt{K\ln K/B})$.
\end{proposition}
% \begin{proof}[Proof Sketch of Proposition \ref{prop:bandit_upper_bound}]
%     The main difference in the proof from previous ones is that, here we directly bounding switching costs by the number of batches instead of using any property from SD.
% \end{proof}
% \duo{I guess the proof sketch and the paragraph below can be removed?}
% In fact, this algorithm can be viewed as a generalization of the original batched EXP3 \citep{Dekel2012OnlineBL} with a flexible batch size $\tau$ and a different way of observing feedback (that is, we only observe one bandit feedback across a batch). One main difference of it from previous algorithms in this paper is that, we are no longer using SD (now $I_{\mathrm{SD}}=0$) to achieve low switching costs. This is because now the budget is relatively small ($B=O(\PTP)$), and thus bounding switching costs by the number of batches $N$ is sufficient to obtain desired result.

% For sure, our unbiased loss estimate differs a little bit, following the standard construction in adversarial MAB.

\begin{remark}
% The above result is encouraging, in the sense that when $B=O(\PTP)$, even using pure bandit feedback can achieve the optimal regret of $\widetilde{\Theta}(T\sqrt{K/B})$. 
% This result also offers an answer to question \textbf{(Q1)} raised in \cref{sec:extra}: not only does it capture the regret scaling with respect to the amount of bandit feedback (i.e., still $\Theta(1/\sqrt{B})$) when $B$ is relatively small, but it also implies that to achieve a regret of  $\wtTheta(\PTP)$, it suffices to use bandit feedback from only $B= \wtTheta(\PTP)$ rounds rather than all $T$ rounds as in the classic algorithms~\citep{Dekel2012OnlineBL}.

This result is encouraging, in the sense that when $B=O(\PTP)$, even using pure bandit feedback can achieve the optimal minimax regret of $\widetilde{\Theta}(T\sqrt{K/B})$. 
This result also answers question \textbf{(Q1)} raised in \cref{sec:extra}. First, it captures the regret scaling with respect to the amount of bandit feedback (i.e., still $\Theta(1/\sqrt{B})$) when $B$ is relatively small. Second, it implies that to achieve a regret of $\wtTheta(\PTP)$, it suffices to use bandit feedback from only $B= \Theta(\PTP)$ rounds rather than all $T$ rounds as in the classic algorithms~\citep{Dekel2012OnlineBL}. The same minimax regret at these two endpoints ($B= \Theta(\PTP)$ and $B=T$) further implies that if only bandit feedback is allowed, the minimax regret is also $\wtTheta(\PTP)$ when $B=\Omega(\PTP)$ (i.e., in-between the two endpoints). 
%\bo{Done. Please check.}
%When $B=\Omega(\PTP)$, 
In this case, bandit feedback is \emph{no longer sufficient} to achieve the optimal minimax regret of $\widetilde{\Theta}(T\sqrt{K/B})$,
% Specifically, it has been shown in \citet{dekel2014bandits} that when only pure bandit feedback is used, the minimax regret is lower bounded by $\wtOmega(\PTP)$ even if one uses bandit feedback from all $T$ rounds (i.e., bandit learning with switching costs), let alone from $B=\Omega(\PTP)$ rounds.
%On the other hand, 
although full-information and flexible feedback can still achieve this optimal minimax regret (see Propositions~\ref{prop:uniform_alg_upper_bound} and \ref{prop:flex upper bound}). 
Clearly, this shows the crucial impact of different types of feedback (when the total budget $B$ is large), which answers the second part of question \textbf{(Q3)}.
%of $\widetilde{\Theta}(T\sqrt{K/B})$.
%
On the other hand, however, a straightforward result (Proposition~\ref{prop:bandit_no_sc_reg} in Appendix~\ref{app:auxiliary}), along with Propositions~\ref{prop:uniform_alg_upper_bound} and \ref{prop:flex upper bound} and Lemma~\ref{lem:total_lower}, shows that in the standard setting without switching costs, all three types of feedback can achieve optimal regret in the full range of $B$. This reveals that the impact of feedback type is partly due to switching costs. 
%\duo{Moreover, with a simple auxiliary proof in Appendix \ref{app:auxiliary}, we can see that without switching costs, all three types can achieve optimal regret in the full range.}
We also summarize these results in Table~\ref{tab:regret_of_different_types}. 
%\duo{By comparing the 2nd and 3rd column of it, we can further conclude that while introducing switching costs does not change the minimax regret, it does make bandit feedback suboptimal when $B$ is large and hence the type of feedback becomes crucial.}
%\bo{done; please check}\duo{good to me. it's quite long, do we need to make it a remark or just a paragraph(main text)?}\bo{I think remark is better}
% This indicates that for a large $B$, one has to go beyond bandit feedback and carefully choose the type of feedback to achieve the optimal scaling of $\Theta(1/\sqrt{B})$, e.g., via algorithms $\pi_{\mathrm{full}}$ and $\pi_{\mathrm{flex}}$ presented in Section~\ref{sec:full and flex alg}. 
%\bo{to be finished}
\end{remark}
\begin{remark}\label{remark:improved_bandit}
    Under bandit feedback, adopting a different regularizer called \emph{Tsallis entropy}~\citep{Audibert2009MinimaxPF} to the OMD framework could further remove the $\sqrt{\ln K}$ factor in the upper bound from Proposition~\ref{prop:bandit_upper_bound} and exactly match the lower bound (order-wise) presented in Lemma~\ref{lem:total_lower}.
    %However, the $\sqrt{\ln K}$ factor is unavoidable with full-information/flexible feedback due to the fundamental lower bound~\citep{Audibert2010RegretBA}.\duo{not confident about flex.}}
\end{remark}

%% file: related.tex
\section{Related Work}\label{sec:related_work}

%\bo{This section is long. Shall we move Related Work to right before Conclusion so that we can introduce the problem setup earlier?}
%\bo{Also, add a sentence to introduce this section?}\xingyu{agree. put related work later}

In this section, we present detailed discussions on several lines of research that are most relevant to ours. We omit the discussion on bandit and expert problems with switching costs as we have discussed this line of work in Section~\ref{sec:intro}.

\textbf{Online Learning with Total Observation Budget.}
In this line of research, the focus is on regret minimization when feedback is not always available and hence ``limited'' within a total budget. For example, in the so-called ``label efficient (bandit) game'' \citep{CesaBianchi2004MinimizingRW, Audibert2010RegretBA}, the learner can ask for full-information/bandit feedback from no more than $m\in[1, T]$ round(s). It is shown that the tight optimal regrets are $\Theta(T\sqrt{\ln{K}/{m}})$ and $\Theta(T\sqrt{K/m})$ under full-information and bandit feedback, respectively.
\citet{seldin2014prediction} also considers a total observation budget in online learning, where the learner can freely request feedback, as long as the total amount of observed losses does not exceed the given total budget $B$.
%{For example, bandit feedback from one round counts for one, and full-information feedback counts for $K$. Note that the learner is not limited to these two specific types of feedback.}
They establish a tight characterization of the minimax regret in their setting (i.e., $\wtTheta(T\sqrt{K/B})$). However, they do not consider switching costs, nor the case when the total observation budget is smaller than $T$ in their algorithm design. %With this key difference, we need to take switching costs into account in both lower bound analysis as well as the algorithm design on top of their work.
Interestingly, we show that introducing switching costs \emph{does not increase} the intrinsic difficulty of online learning in the sense that the minimax regret remains $\wtTheta(T\sqrt{K/B})$, but the feedback type becomes crucial.

\textbf{Bandits with Additional Observations.} \citet{Yun2018MultiarmedBW} considers the bandit setting with additional observations, where the learner can freely make $n\in[0,K-1]$ observations at each round in addition to the bandit feedback. Hence, this can be viewed as a special case of online learning with a total observation budget~\citep{seldin2014prediction}. That is, a total of $(n+1)T$ observations are used in a particular way (i.e., bandit plus extra observations). They present a tight characterization of the scaling of the minimax regret with respect to $K$, $T$, and $n$. Similar to \citet{seldin2014prediction}, however, switching costs are not considered.

\textbf{Online Learning with Switching Costs and Feedback Graphs.}
\citet{arora2019bandits} considers online learning with switching costs and feedback graphs, where given a feedback graph $G$, the learner observes the loss associated with the neighboring action(s) of the chosen action (including itself).
However, the feedback graph is given and hence the additional feedback is \emph{not} of the learner's choice. \citet{arora2019bandits} shows that in this setting, the minimax regret is $\wtTheta(\gamma(G)^{1/3} T^{2/3})$, where $\gamma(G)$ is the domination number of the feedback graph $G$. 
Hence, the dependency on $T$ remains the same as in the standard bandit setting without additional observations (i.e., $\wtTheta(T^{2/3})$).
%In the worst case, one still cannot improve the regret with respect to $T$, compared to the bandit setting. 
On the contrary, in the setting we consider, the learner can freely decide the loss of which actions to observe, which leads to different (and more interesting) regret bounds.

\textbf{Online Learning with Limited Switches.}
\citet{altschuler2018online} considers online learning with limited switches. In contrast to the settings with switching costs, here the learner does not pay additional losses for switching actions; instead, the total number of switches allowed is capped at $S$. Compared to our setting, a key difference is that switching is a constraint rather than a penalty added to the loss/cost function. 
%\citet{altschuler2018online} exploits this setting by considering that $S$ is a hard cap on total number of switches. 
They show that in the bandit setting, the minimax regret is $\widetilde{\Theta}(T\sqrt{K/S})$, i.e., the regret improves as the switching budget increases;
in the expert setting, however, there is a phase-transition phenomenon: while the minimax regret is $\wtTheta(T \ln{K} / S)$ when $S=O(\sqrt{T \ln{K}})$, it remains $\wtTheta(\sqrt{T \ln{K}})$ when $S=\Omega(\sqrt{T \ln{K}})$.
%\bo{I think it is fine to use $S$ here as it is also the number of switches}\duo{I think you are right}
%\bo{Shouldn't the range of $S$ be switched? Can you double check?}\duo{when $S$ is large, it doesn't affect the regret, otherwise it increases the regret, right? I think it's correct after checking the 2018 COLT paper.}\bo{thanks for checking.}

\textbf{Online Learning against Adaptive Adversaries}. Online learning with switching costs can also be viewed as a special case of \emph{learning against adaptive adversaries}, where the losses at round $t$ are adapted to actions taken at both rounds $t$ and $t-1$ (in contrast to the oblivious adversaries we consider). Such adversaries have a \emph{bounded memory} (of size one), in the sense that they could adapt only up to the \emph{most recent} action, instead of any history in the earlier rounds~\citep{cesa2013online}. Adopting the multi-scale random walk argument in~\citet{dekel2014bandits}, it has been shown that against \emph{adaptive adversaries with a memory of size one}, the \emph{minimax policy regret} is $\wtTheta(T^{2/3})$ under \emph{both} bandit feedback~\citep{cesa2013online} and full-information feedback~\citep{feng2018online}. This is fundamentally different from the special case with switching costs, where the minimax regret is different under bandit feedback and full-information feedback ($\wtTheta(T^{2/3})$ vs. $\wtTheta(\sqrt{T})$).
%\bo{try to eliminate lines with widow words through rephrasing}\xingyu{like this suggestion, which is also what I aim for sometimes:)}

\paragraph{Stochastic Bandits and the Best of Both Worlds.}Note that the above discussions have been focused on the adversarial setting. There is another body of work focused on the stochastic setting (see, e.g., \citet{Auer2002FinitetimeAO,Auer2003UsingCB, SimchiLevi2019PhaseTI}), where the loss/reward follows some fixed distribution rather than being generated arbitrarily by an adversary. Hence, it is very different from the adversarial setting we consider. An interesting line of work has been focused on designing algorithms that can perform well in both adversarial and stochastic settings, thus achieving \emph{the best of both worlds} (see, e.g., \citet{Bubeck2012TheBO, Zimmert2019BeatingSA}).

\paragraph{Other Related Work.}
In \citet{shi2022power}, a novel bandit setting with switching costs and additional feedback has been considered.
%First, we elaborate more on \citet{shi2022power} to show the difference between their work and ours. 
Specifically, the learner maintains an ``action buffer'' for each round, which is a subset of actions with fixed cardinality $m\in[K]$, and the learner can only take an action from this buffer set. Their switching cost can be roughly viewed as how much change is made to this buffer set throughout the game -- replacing an action in the buffer set incurs a constant cost. While the learner can observe the losses of all the actions in this buffer set for free, the learner can also choose to receive full-information feedback (i.e., observing the losses of all actions rather than just actions in the buffer set) by paying another (larger) constant cost. 
Although we draw inspiration from their work 
for deriving the lower bound and designing algorithms, both their problem setup and regret definition are very different from ours, and more importantly, they do not consider observation budget.

%% file: app.tex
\newpage

\appendix
\onecolumn

% Among all the discussions on related work above, we focus on adversarial rewards, while there are also many relevant works in the stochastic world \citep{Auer2002FinitetimeAO}.
% \citet{Yun2018MultiarmedBW} also proposes algorithms to make use of additional observations in stochastic bandits (without switching costs).
% \citet{SimchiLevi2019PhaseTI} provides a comprehensive and tight characterization on the minimax regret in stochastic bandits with limited switches.
% \citet{amir2022better} studies best of both words algorithms in online learning with switching costs. In particular, the environment could either be adversarial or \emph{stochastically-constrained}. The goal is to achieve desirable regret guarantee without being aware of the environment type.

\section{Motivating Examples for Online Learning with Switching Costs and Observation Budget}\label{app:motivate}
%\duo{Another option: in network configuration problem, we want to adapt to the optimal configuration (action) in an online manner. First, changing configuration takes some cost so it's natural to take switching costs into account. Then, after we make the decision (configuration) without being aware of the underlying network condition and then receives feedback, we become aware of the network condition, and hence we can run some simulations to see what if we take other actions, which could reveal the losses of other actions in the same round. However, due to computation resource limitation, we may want to have a limited budget on the total amount of simulations (additional extra observation)}

Consider a retail company that uses online learning to improve its website user interface (UI) design in order to attract more users. In this case, actions correspond to different UI designs. First, switching costs should be taken into account as frequently changing the website interface may become annoying to users. To evaluate other actions (different UI designs), the company can run an A/B test and display different interfaces to separate and relatively small groups of users so that the feedback of other actions is also available (in addition to the one displayed to the main and large population). However, each different website needs additional resources to be deployed and maintained, and hence, one may want to impose a total observation budget.
%\xingyu{this is better than networking one. use this one first, which is easier for people to understand. For the edge one below, we can modify it. we do not have to consider edge, just some ML pipeline applications. or we can use the above networking configuration one. I am fine with this part.}\bo{yes, this one is easier to understand; we can use this one first - it is always good to have more than one, but we need to make the second one clear}

Another example would be Machine Learning as a Service (MLaaS). Consider a company that uses large ML models for jobs like prediction, chatbots, etc. They may train several different models and dynamically choose the best one via online learning. Changing the deployed ML model is not free: the new model needs to be loaded (which could be costly, especially nowadays when the number of parameters is quite large), 
and other components in the pipeline may also need to be adjusted accordingly. As a result, it is natural that redeploying or updating model components would incur a cost. While the performance of the deployed model is easily accessible, the company can also run these jobs using other models that are not deployed in the production system, to receive additional feedback. However, running these jobs consumes additional resources (e.g., computing and energy), which is not free either. Therefore, one may want to impose a budget on the number of additional observations (i.e., evaluations).
%\xingyu{"sample" is not clear}\duo{test sample?}\duo{data?}\xingyu{just say processing task. classification, text generation, and translation..., i.e., complete these tasks using ML models...}
 
%A motivating example for observation budget we consider comes from an application scenario in edge computing~\citep{Murshed2019MachineLA, shi2022power}. Suppose that an image is delivered to an edge server for processing (e.g., classification \citep{He2015DelvingDI} and denoising \citep{Buades2005ARO}). There are multiple Machine Learning (ML) models that one can choose to use, which, however, are stored in a remote cloud and cannot be simultaneously deployed at the edge server due to its limited storage capacity. Each time after the edge server finishes processing the image using the chosen ML model, the performance (e.g., accuracy) of this model is revealed. Additionally, we can also evaluate the performance of the other ML models for the same image. However, these models need to be transmitted from the cloud to the edge server for evaluations, 
%through wireless channels, thus incurring additional latency or computation overhead~\citep{Elbamby2019WirelessEC}. Therefore, even if one can make extra observations at her own choice, there must be a limit on the total number of such observations to avoid significant overhead.
%\bo{This does not seem to be a good motivating example.}

% \xingyu{remove all unnecessary equation numbers?}\duo{done}

\section{Proof of Proposition \ref{prop:bandit_extra_lower_bound}}\label{app:extra_lower_bound_proof}
\begin{proof}[Proof of Proposition \ref{prop:bandit_extra_lower_bound}]
Our proof is built on Yao's minimax principle \cite{yao1977crobabilistic}. That is, we will establish a lower bound on the expected regret for any deterministic algorithm over a class of randomly generated loss sequences, which further implies the same lower bound for any randomized algorithm over some loss sequence in this class.

To begin with, we would like to give some high-level ideas about the proof.
Note that while the loss sequence generation method we adopt will be the same as Algorithm~1 in \citet{shi2022power}, we need a different analysis to establish the lower bound due to a different setting we consider. Specifically, in the original loss sequence construction based on multi-scale random walk~\citep{dekel2014bandits}, the optimal action $k^*$ has the lowest loss at all $T$ rounds. With bandit feedback, useful information toward distinguishability (between the optimal action and suboptimal
actions) is gained only when the learner switches between actions. With full-information feedback, however, the learner can immediately identify the optimal action even at one round only. Therefore, to construct a hard instance (i.e., loss sequence) for the setting where the learner is equipped with additional observations beyond the bandit feedback, \citet{shi2022power} introduced an action-dependent noise in addition to the original loss (which is called the \emph{hidden loss}). Now, the learner's information comes from two parts. On the one hand, the learner still gains distinguishability from switches (which is related to hidden losses). On the other hand, conditioning on hidden losses, the extra observations also provide additional information. Combining two parts together, we obtain a lower bound related to both the number of switches and the number of extra observations. For convenience, we restate this loss sequence generation method in Algorithm~\ref{alg:purdue_instance}. Specifically, we first generate the sequence $\{G(t)\}_t$ according to the random walk design (Line~\ref{line:alg2_random_walk}). Next, we determine the loss before truncation, i.e., $\ell_t^{\text{init}}[k]$ (Line~\ref{line:alg2_ell_init}). We first add an action-dependent noise $\gamma_k(t)$ (which is an \emph{i.i.d.} Gaussian random variable) to $G(t)$ for each action $k\in[K]$. And then, for the optimal action $k^*$ only, we will further subtract $\epsilon$ (which is determined in the very beginning as an input to the algorithm) from the value obtained after adding $\gamma_k(t)$. Finally, we truncate each $\ell_t^{\text{init}}[k]$ onto range $[0,1]$ and obtain $\ell_t[k]$ (Line~\ref{line:alg2_ell}).
%, where $\mathcal{N}(0,\sigma^2)$ denotes Gaussian distribution with mean zero and variance $\sigma^2$.}

\begin{algorithm}[t]
    \caption{Loss Sequence Generation Method~\citep{shi2022power}}\label{alg:purdue_instance}
    \begin{algorithmic}[1]
        \STATE \textbf{Input:} suboptimality gap $\epsilon$ and noise variance $\sigma^2$
        \STATE \textbf{Initialization:} choose the identity of optimal action $k^*$ uniformly at random from $[K]$; initialize Gaussian process $G(t)=0, \forall t\geq 0$; define functions $\delta(t):=\max\{i\geq 0:2^i \text{~divides~}t\}$ and  $\rho(t):=t-2^{\delta(t)}, \forall t\geq 0$
        
        \FOR {time $t = 1: T$}
            \STATE $G(t)=G(\rho(t))+\xi(t)$, where each $\xi(t)$ is an \emph{i.i.d.} sample from $\mathcal{N}(0,\sigma^2)$\alglinelabel{line:alg2_random_walk}
            \STATE $\ell_t^{\mathrm{init}}[k] = G(t)-\epsilon\cdot \bI_{\{k=k^*\}}+\gamma_k(t),\forall k \in[K]$, where $\gamma_k(t)$ is an \emph{i.i.d.} sample from Gaussian distribution $\mathcal{N}(0,\sigma^2)$\alglinelabel{line:alg2_ell_init}
            \STATE $\ell_t[k] = \arg\min_{z\in[0,1]}|z-\ell_t^{\mathrm{init}}[k]|, \forall k\in[K]$\alglinelabel{line:alg2_ell} %\bo{you earlier used $x$ to denote an action}
        \ENDFOR
    \end{algorithmic}
\end{algorithm}

Next, we give some additional notations needed for this proof. For any $k\in[K]$, let $\bP_{k}$ denote the conditional probability measure given the special (i.e., optimal) action $k^{*}=k$, i.e., $\bP_{k}(\cdot):= \bP(\cdot |k^{*}=k)$. As a special case, when $k^*=0$,  $\bP_{0}$ denotes the conditional probability measure where all the actions are identical and there is no special action. Let $\bE_{k}[\cdot]:=\bE[\cdot|k^*=k]$ denote the conditional expectation under measure $\bP_{k}$, and let $\bE[\cdot]:=\frac{1}{K}\sum_{k=1}^K\bE_{k}[\cdot]$. Let $\ell^{\mathrm{ob}}_{1:T}$ denote the observed loss sequence throughout the game. For two probability distributions $\mathcal{P}$ and $\mathcal{Q}$ on the same space, let $D_{\mathrm{KL}}(\mathcal{P}\|\mathcal{Q}):=\bE_{x\sim \mathcal{P}}[\log \left(d\mathcal{P}(x)/d\mathcal{Q}(x)\right)]$ denote the Kullback-Leibler (KL) divergence (i.e., relative
entropy) between $\mathcal{P}$ and $\mathcal{Q}$, and let $D_{\mathrm{TV}}(\mathcal{P}\|\mathcal{Q}):=\sup\{\mathcal{P}(A)-\mathcal{Q}(A):A \text{~measurable}\}$ denote the total variation distance between $\mathcal{P}$ and $\mathcal{Q}$.

Let $S_t^{k}:= \bI_{\{X_{\rho(t)} = k, X_t \neq k\}} + \bI_{\{X_{\rho(t)} \neq k, X_t = k\}}$ be the indicator of whether it is switched from \emph{or} to action $k$ between rounds $\rho(t)$ and $t$, let $\bar{S}^{k} := \sum_{t=1}^T S_t^{k}$ be the total number of action switches from \emph{or} to action $k$, let $\bar{S}$ be the total number of action switches, i.e., $\bar{S}:=\sum_{t=1}^T\bI_{\{X_t\neq X_{t-1}\}}=\sum_{k=1}^K \bar{S}^{k}/2$, and let $N_{\mathrm{ex}}^t$ be the number of extra observations made at round $t$ in addition to the bandit feedback. Then, we naturally have $N_{\mathrm{ex}}^t\in[0,K-1]$ and $\sum_{t=1}^T N_{\mathrm{ex}}^t \le \Bex$ since the learning algorithm makes no more than $\Bex$ extra observations in total. Let $R_T$ be the regret of the deterministic learning algorithm interacting with the loss sequence $\ell_{1:T}$, and let $R_T^{\mathrm{init}}$ be the (hypothetical) regret on \emph{the same action sequence} with respect to loss sequence $\ell_{1:T}^{\text{init}}$.

%According to the Appendix B in \citet{shi2022power}, 
In the following proof, we need Lemmas~A.1 and A.4 of \citet{shi2022power} and Lemma~2 of \citet{dekel2014bandits}. We restate these three results 
%in Appendix~\ref{app:auxiliary} 
as Lemmas~\ref{lem:shi_lem_restate}, \ref{lem:dekel_lem_gap_restate}, and~\ref{lem:dekel_lem_width_restate}, respectively.

%\bo{I feel it would be better to state these lemmas before using them. However, the way they are written is just a list of results. Some more explanations may be needed, and the transition needs to be made more smooth.}

\begin{lemma}\citep[Lemma~A.1 (restated)]{shi2022power}\label{lem:shi_lem_restate}
    The KL divergence between  $\bP_0\left(\ell^{\mathrm{ob}}_{1:T}\right)$ and $\bP_{k}\left(\ell^{\mathrm{ob}}_{1:T}\right)$ can be upper bounded as follows:
    \begin{align}
        &D_{\mathrm{KL}}\left(\bP_0\left(\ell^{\mathrm{ob}}_{1:T}\right) \| \bP_{k}\left(\ell^{\mathrm{ob}}_{1:T}\right)\right) \nonumber\\
&\leq \sum_{t=1}^T\left[\bP_0\left(N_{\mathrm{ex}}^t=0, S_t^{k}=1\right) \cdot \frac{\epsilon^2}{2 \sigma^2}+ \sum_{j=1}^{K-1} \bP_0 \left(N_{\mathrm{ex}}^t=j, S_t^{k}=0, X_t \neq k\right) \cdot \frac{\epsilon^2}{2 \sigma^2}\right. \nonumber\\
& \quad + \sum_{j=1}^{K-1} \bP_0 \left(N_{\mathrm{ex}}^t=j, S_t^{k}=0, X_t=k\right) \cdot \frac{j \epsilon^2}{2 \sigma^2}+ \sum_{j=1}^{K-1} \bP_0 \left(N_{\mathrm{ex}}^t=j, S_t^{k}=1, X_t \neq k\right) \cdot \frac{2\epsilon^2}{2 \sigma^2} \nonumber \\
& \quad \left. + ~ \sum_{j=1}^{K-1} \bP_0\left(N_{\mathrm{ex}}^t=j, S_t^{k}=1, X_t=k\right) \cdot \frac{(j+1) \epsilon^2}{2 \sigma^2}\right].\nonumber
    \end{align}
\end{lemma}
Lemma~\ref{lem:shi_lem_restate} is obtained by considering five disjoint cases (which corresponds to the five terms on the Right-Hand-Side (RHS) in terms of different values of $N_{\text{ex}}^t$, $S_t^k$, and $X_t$. This lemma reveals the relationship between the KL divergence and the number of switches and the number of extra observations and will be used for deriving Eq.~\eqref{eqn:bound_KL}.

\begin{lemma}\citep[Lemma~A.4 (restated)]{shi2022power}\label{lem:dekel_lem_gap_restate}
    Consider the instance construction in Algorithm~\ref{alg:purdue_instance}. Suppose that we have $\epsilon\leq 1/6$ and $\sigma = 1/(9\log_2 T)$. Then, the difference between $\ex{R_T}$ and $\ex{R_T^{\mathrm{init}}}$ can be bounded as follows:
    $$\ex{R_T^{\mathrm{init}}}-\ex{R_T}\leq \frac{\epsilon T}{6}.$$
\end{lemma}
%\bo{The above statement is strange. You consider one instance construction but use a result based on a different instance construction.}\duo{fixed}
%\bo{$R_T^{\mathrm{init}}$ and $R_T$ have not been defined yet. You really need to come back and read the entire proof again after you made edits, pretending that you are a new reader.}\duo{added above Lemma~2}
Although the multi-scale random walk serves as a powerful and convenient tool to help us obtain the desired lower bound, an issue is that the losses could lie out of the range $[0,1]$, which does not satisfy our problem setup. That is, based on the random walk, we can derive a lower bound on $\ex{R_T^{\text{init}}}$, with respect to a possibly unbounded loss sequence $\ell_{1:T}^{\text{init}}$. 
%\duo{in this case, I think it would be better to put these three lemmas in the last section to avoid undefined notations, as some relevant details are not introduced until later.  Otherwise we need to move them above here.} \bo{I still prefer to state these lemmas here. You will have the same issue even if you move them to the last section as this is the point you mention the restatements.}
However, our goal is to obtain a lower bound with respect to the bounded losses. To get around this issue, Lemma~\ref{lem:dekel_lem_gap_restate} presents a useful result: if $\epsilon$ and $\sigma$ satisfy certain conditions, then the difference between $\ex{R_T^{\text{init}}}$ and $\ex{R_T}$ will not be too large, which is sufficient to give us the desired result.
%\bo{Duo, you really need to use some grammar-checking tools (e.g., Grammarly or ChatGPT) to help you check the basics throughout the paper. I use Grammarly integrated with overleaf, which quickly spots several such issues.}

\begin{lemma}\citep[Lemma~2 (restated)]{dekel2014bandits}\label{lem:dekel_lem_width_restate}
Under the instance construction in Algorithm~\ref{alg:purdue_instance}, the following is satisfied:
$$\sum_{t=1}^T\bP_0\left(S_t^{k}=1\right) = \ex{\sum_{t=1}^T S_t^k}\leq (\lfloor \log_2 T\rfloor +1) \cdot \bE_{0}\left[\bar{S}^{k}\right].$$
Furthermore, it can be bounded by $2\log_2 T \cdot \bE_{0}\left[\bar{S}^{k}\right]$ for a sufficiently large $T$.
\end{lemma}

\begin{remark}
Lemma~\ref{lem:dekel_lem_width_restate} holds regardless of whether the action-dependent is added or not. Therefore, it is true under the instance constructions from both \citet{dekel2014bandits} and \citet{shi2022power}.
\end{remark}

Lemma~\ref{lem:dekel_lem_width_restate} relies on careful design of the random walk. We refer interested readers to \citep[Section~3]{dekel2014bandits} for technical details.
%, especially regarding the ``depth'' and ``width''. 
In the following proof, we will first bound the KL divergence in part by $\bE[\sum_{t=1}^T S_t^k]$. This term is different from the switching costs we consider, as $S_t^k$ roughly denotes the switch between rounds $\rho(t)$ and $t$, \emph{rather than between two consecutive rounds}. To handle this difference, Lemma~\ref{lem:dekel_lem_width_restate} builds a connection between the two and will be used for deriving Eq.~\eqref{eqn:bound_KL}.

With the above three restated lemmas, we are now ready to derive an upper bound on the KL divergence between $\bP_0\left(\ell^{\mathrm{ob}}_{1:T}\right)$ and $\bP_{k}\left(\ell^{\mathrm{ob}}_{1:T}\right)$. In particular, for any $k \in [K]$, we have
\begin{align}
&D_{\mathrm{KL}}\left(\bP_0\left(\ell^{\mathrm{ob}}_{1:T}\right) \| \bP_{k}\left(\ell^{\mathrm{ob}}_{1:T}\right)\right) \nonumber\\
&\overset{\text{(a)}}{\leq} \sum_{t=1}^T\left[\bP_0\left(N_{\mathrm{ex}}^t=0, S_t^{k}=1\right) \cdot \frac{\epsilon^2}{2 \sigma^2}+ \sum_{j=1}^{K-1} \bP_0 \left(N_{\mathrm{ex}}^t=j, S_t^{k}=0, X_t \neq k\right) \cdot \frac{\epsilon^2}{2 \sigma^2}\right. \nonumber\\
& \quad + \sum_{j=1}^{K-1} \bP_0 \left(N_{\mathrm{ex}}^t=j, S_t^{k}=0, X_t=k\right) \cdot \frac{j \epsilon^2}{2 \sigma^2}+ \sum_{j=1}^{K-1} \bP_0 \left(N_{\mathrm{ex}}^t=j, S_t^{k}=1, X_t \neq k\right) \cdot \frac{2\epsilon^2}{2 \sigma^2} \nonumber \\
& \quad \left. + ~ \sum_{j=1}^{K-1} \bP_0\left(N_{\mathrm{ex}}^t=j, S_t^{k}=1, X_t=k\right) \cdot \frac{(j+1) \epsilon^2}{2 \sigma^2}\right]\nonumber  \\
&\overset{\text{(b)}}{\leq} \sum_{t=1}^T\left[\bP_0\left(N_\mathrm{ex}^t=0, S_t^{k}=1\right) \cdot \frac{\epsilon^2}{2 \sigma^2}+ \sum_{j=1}^{K-1}\frac{j\epsilon^2}{\sigma^2}\cdot \left( \bP_0 \left(N_{\mathrm{ex}}^t=j, S_t^{k}=0, X_t \neq k\right) + \bP_0 \left(N_{\mathrm{ex}}^t=j, S_t^{k}=0, X_t=k\right) \right. \right. \nonumber\\
& \quad \left.\left. + \bP_0 \left(N_{\mathrm{ex}}^t=j, S_t^{k}=1, X_t \neq k\right) + \bP_0\left(N_{\mathrm{ex}}^t=j, S_t^{k}=1, X_t=k\right)\right)\right]\nonumber \\
&\overset{\text{(c)}}{\leq} \sum_{t=1}^T\left[\bP_0\left(S_t^{k}=1\right) \cdot \frac{\epsilon^2}{2 \sigma^2} + \sum_{j=1}^{K-1} \bP_0 \left(N_{\mathrm{ex}}^t=j\right) \cdot \frac{j\epsilon^2}{ \sigma^2} \right]\nonumber \\
&\overset{\text{(d)}}{\leq} 
2\log_2 T \cdot \frac{\epsilon^2}{2\sigma^2} \cdot \left( \bE_{0}\left[\bar{S}^{k}\right] + 2 \Bex\right), \label{eqn:bound_KL}
\end{align}
where (a) is from Lemma~\ref{lem:shi_lem_restate}, (b) is obtained by enlarging the last four terms using the fact that $2 \leq j+1 \leq 2j, \forall j \geq 1$, (c) is obtained by applying the monotonicity property of probability to the first term and merging the last four disjoint events,
and (d) is from Lemma~\ref{lem:dekel_lem_width_restate} and the fact that $\sum_{t=1}^T \sum_{j=1}^{K-1} \bP_0(N_{\mathrm{ex}}^t = j)\cdot j= \sum_{t=1}^T \bE_{0}\left[N_{\mathrm{ex}}^t\right] \le \Bex$. 
%\bo{I feel the above argument is not self-contained. Is it possible to restate those lemmas from the references?}\duo{fixed}
Note that Eq.~(\ref{eqn:bound_KL}) indicates that the KL divergence (which can be viewed as the information obtained by the learner) is related to both the number of switches and the amount of extra feedback.
%, which is consistent with the high-level idea we give at the beginning of this section.} 
With the upper bound on the KL divergence in Eq.~(\ref{eqn:bound_KL}), we can also bound the total variation. Specifically, we have %\bo{use \eqref{} instead of (\ref{}) for equation reference in the future; do not bother changing it this time}
\begin{align}
&\frac{1}{K} \sum_{k=1}^K D_{\mathrm{TV}} \left(\bP_0\left(\ell^{\mathrm{ob}}_{1:T}\right) \| \bP_{k}\left(\ell^{\mathrm{ob}}_{1:T}\right)\right) \nonumber \\
&\overset{\text{(a)}}{\leq} \frac{1}{K} \sum_{k = 1}^K \sqrt{\frac{\ln 2}{2}} \sqrt{D_{\mathrm{KL}}\left(\bP_0\left(\ell^{\mathrm{ob}}_{1:T}\right)\nonumber \| \bP_{k}\left(\ell^{\mathrm{ob}}_{1:T}\right)\right)}\\
&\overset{\text{(b)}}{\leq} \sqrt{\frac{\ln 2}{2}}\sqrt{2\log _2 T \cdot \frac{\epsilon^2}{2\sigma^2}} \cdot \frac{1}{K} \sum_{k = 1}^K \sqrt{\bE_{0}\left[\bar{S}^{k}\right] + 2  \Bex}\nonumber \\
&\overset{\text{(c)}}{\leq} \sqrt{\frac{\ln 2}{2}}\sqrt{2\log _2 T \cdot \frac{\epsilon^2}{2\sigma^2}} \cdot \sqrt{\frac{1}{K} \sum_{k = 1}^K   \left( \bE_{0}\left[\bar{S}^{k}\right] + 2  \Bex \right)}\nonumber  \\
&\overset{\text{(d)}}{=} \frac{\epsilon}{\sigma}\sqrt{\frac{\ln 2\cdot \log _2 T}{K}} \sqrt{ \bE_{0}\left[\bar{S}\right] + \Bex }\label{eqn:bound_TV},
\end{align}
where (a) is from Pinsker's inequality, (b) is from Eq.~(\ref{eqn:bound_KL}), (c) is from Jensen's inequality, and (d) is from $\sum_{k = 1}^K \bE_{0}\left[\bar{S}^{k}\right] = 2\bE_{0}\left[\bar{S}\right]$.

With all the above results, we are ready to derive a lower bound on $\ex{R_T^{\mathrm{init}}}$ after showing two intermediate steps. Let $N_k$ be the number of times action $k\in [K]$ is played up to round $T$ (which is a random variable). We first assume that the deterministic learning algorithm makes at most $\epsilon T$ switches on \emph{any} loss sequence, which will be used for deriving Eq.~\eqref{eqn:switch_TV} below, and we will later relax this assumption. Under this assumption, we have
\begin{align}
\bE_{0}\left[\bar{S}\right] - \bE\left[\bar{S}\right] &\overset{\text{(a)}}{=} \frac{1}{K} \sum_{k=1}^{K}\left(\bE_{0}\left[\bar{S}\right] - \bE_{k}\left[\bar{S}\right]\right)\nonumber \\
& \overset{\text{(b)}}{=} \frac{1}{K} \sum_{k=1}^{K}\sum_{z=1}^{T}\left(\bP_0(\Bar{S}\geq z) - \bP_k(\Bar{S}\geq z) \right)\nonumber \\
& \overset{\text{(c)}}{=} \frac{1}{K} \sum_{k=1}^{K}\sum_{z=1}^{\epsilon T}\left(\bP_0(\Bar{S}\geq z) - \bP_k(\Bar{S}\geq z) \right)\nonumber \\
&\overset{\text{(d)}}{\leq} \frac{\epsilon T}{K} \sum_{k=1}^K D_{\mathrm{TV}} \left(\bP_0\left(\ell^{\mathrm{ob}}_{1:T}\right) \| \bP_{k}\left(\ell^{\mathrm{ob}}_{1:T}\right)\right)\label{eqn:switch_TV},
\end{align}
where (a) is from the definition that $\bE[\cdot]:=\frac{1}{K}\sum_{k=1}^K\bE_{k}[\cdot]$, (b) is from rewriting the expectations, (c) is from the assumption of no more than $\epsilon T$ switches, and (d) is from the definition of the total variation. Also, we have
\begin{align}
 \sum_{k=1}^K \bE_{k}[N_k] - T&\overset{\text{(a)}}{=} \sum_{k=1}^K \left( \bE_{k}[N_k] - \bE_{0}[N_k]\right)\nonumber\\
& \overset{\text{(b)}}{=} \sum_{k=1}^{K}\sum_{z=1}^{T}\left(\bP_k(N_k \geq z) - \bP_0(N_k \geq z) \right)\nonumber \\
&\overset{\text{(c)}}{\leq} T\sum_{k=1}^K D_{\mathrm{TV}} \left(\bP_0\left(\ell^{\mathrm{ob}}_{1:T}\right) \| \bP_{k}\left(\ell^{\mathrm{ob}}_{1:T}\right)\right)\label{eqn:pull_TV},
\end{align}
where (a) is from $\sum_{k=1}^K \bE_{0}[N_k] = T$, (b) is from rewriting the expectations, and (c) is from the definition of the total variation.
%and (b) is  \textcolor{blue}{which can be obtained for similar reasons as above}.
%\bo{explain the above}\duo{done}
%\bo{move \& to in front of the equality/inequality; see the changes I made in Eq.~(9)}\duo{have fixed the \&}

We now lower bound the expected value of $R_T^{\mathrm{init}}$ as follows:
\begin{align}
\bE\left[R_T^{\mathrm{init}}\right] &\overset{\text{(a)}}{=} \frac{1}{K} \sum_{k=1}^K \bE \left[\epsilon(T-N_k) + \bar{S}|k^* = k \right]\nonumber \\
&= \epsilon T - \frac{\epsilon}{K}\sum_{k=1}^K \bE_{k}\left[{N}_{k}\right] + \bE \left[\bar{S}\right]\nonumber \\
& \overset{\text{(b)}}{\geq} \epsilon T - \frac{\epsilon T}{K} - \frac{2\epsilon T}{K} \sum_{k=1}^K D_{\mathrm{TV}} \left(\bP_0\left(\ell^{\mathrm{ob}}_{1:T}\right) \| \bP_{k}\left(\ell^{\mathrm{ob}}_{1:T}\right)\right) + \bE_{0} [\bar{S}]\nonumber \\
&\overset{\text{(c)}}{\geq} \frac{\epsilon T}{2} - \frac{2\sqrt{\ln 2} \cdot \epsilon^2 T  \sqrt{\log _2 T} }{\sigma\sqrt{K}}  \sqrt{ \bE_{0}\left[\bar{S}\right] + \Bex} + \bE _{0}\left[\bar{S}\right]\nonumber\\
&\overset{\text{(d)}}{\geq} \frac{\epsilon T}{2} - \frac{2\sqrt{\ln 2} \cdot \epsilon^2 T  \sqrt{\log _2 T} }{\sigma\sqrt{K}} \left( \sqrt{ \bE_{0}\left[\bar{S}\right]} +  \sqrt{ \Bex}\right) + \bE_{0}\left[\bar{S}\right] \nonumber \\
&\overset{\text{(e)}}{\geq} \frac{\epsilon T}{2} - \frac{\ln 2 \cdot   \epsilon^4 T^2  \log_2 T}{K \sigma^2} - \frac{2 \sqrt{ \ln 2} \cdot \epsilon^2 T \sqrt{\log_2 T} }{\sigma \sqrt{K}}  \sqrt{\Bex}, \label{eqn:uni_extra_LB}
\end{align}
where (a) is from the regret definition (i.e., Eq.~\eqref{eqn:reg_def}), (b) is from Eqs.~(\ref{eqn:switch_TV}) and (\ref{eqn:pull_TV}), (c) is from Eq.~(\ref{eqn:bound_TV}), (d) is from an elementary inequality: $ \sqrt{x} + \sqrt{y} \geq \sqrt{x+y}, \forall x,y\geq0$, and (e) is obtained by minimizing the quadratic function of $\sqrt{ \bE_{0}[\bar{S}]}$. 

Now, we turn to lower bound $\ex{R_T}$. By Lemma~\ref{lem:dekel_lem_gap_restate},
if it holds that $\epsilon \leq 1/6$ and $\sigma = 1/(9\log_2 T)$, then we have $\ex{R_T^{\mathrm{init}}}-\ex{R_T}\leq \epsilon T/6$. We first assume $\epsilon \leq 1/6$ and later show that the selected $\epsilon$ satisfies this condition. Then, we have
\begin{align}
\ex{R_T} &\ge \ex{R_T^{\mathrm{init}}} - \epsilon T/6 \nonumber \\
&\geq \frac{\epsilon T}{3} - \frac{81\ln2 \cdot   \epsilon^4 T^2  (\log_2 T)^3}{K } - \frac{ 18\sqrt{\ln 2} \cdot \epsilon^2 T (\log_2 T)^{3/2} }{\sqrt{K}}  \sqrt{  \Bex},
\end{align}
where the last step is from Eq.~(\ref{eqn:uni_extra_LB}) and choosing $\sigma = 1/(9\log_2 T)$.
%\bo{where are the $\ln 2$ and $\sqrt{\ln 2}$ terms in the last step?}\bo{also, why does $9\sqrt{2}$ become 18 in the next step? this is not that important though}\duo{now I keep the original constants, do not change them.}

We now consider two cases for $\Bex$: $\sqrt{\Bex} \leq c_1 {K^{1/6}T^{1/3}}$ and $\sqrt{\Bex} > c_1 {K^{1/6}T^{1/3}}$, for some $c_1>0$.

In the first case of $\sqrt{\Bex} \leq c_1 {K^{1/6}T^{1/3}}$, we have
\begin{align}
\ex{R_T} &\overset{\text{(a)}}{\geq} \frac{\epsilon T}{3} - \frac{81\ln2 \cdot   \epsilon^4 T^2  (\log_2 T)^3}{K } - \frac{18\sqrt{\ln 2}\cdot c_1 \epsilon^2 T (\log_2 T)^{3/2}}{ \sqrt{K}}  \cdot {K^{1/6}T^{1/3}}\nonumber \\
&\overset{\text{(b)}}=  \frac{c_2 K^{1/3}T^{2/3}}{3 (\log_2 T)^{3/2}} - \frac{81\ln2\cdot c_2^4   K^{1/3} T^{2/3}  }{(\log_2 T)^3} - \frac{18\sqrt{\ln 2}\cdot c_1 c_2^2 K^{1/3} T^{2/3} }{(\log_2 T)^{3/2}} \nonumber\\
&\overset{\text{(c)}}{\geq} \left(\frac{c_2}{3} - 81 \ln2 \cdot c_2^4 - 18 \sqrt{\ln 2}\cdot c_1 c_2^2\right) \frac{K^{1/3} T^{2/3}}{18 (\log_2 T)^3}\nonumber \\
&= \widetilde{\Omega}(K^{1/3}T^{2/3})\label{eqn:LB_case_1},
\end{align}
where (a) is from $\sqrt{\Bex} \le c_1 {K^{1/6}T^{1/3}}$, (b) is obtained by choosing $\epsilon = c_2\frac{K^{1/3}}{ T^{1/3} (\log_2 T)^{3/2}}$ (where $c_2>0$ satisfies $\frac{1}{3} - 81\ln 2 \cdot c_2^3 - 18\sqrt{\ln 2}\cdot c_1 c_2 >0$), and (c) is simply due to $(\log_2 T)^3\geq (\log_2 T)^{3/2}$ for a sufficiently large $T$. Since $K\leq T$, we have $\epsilon \leq 1/6$ when $T$ is sufficiently large.
%For the selection of $\epsilon$, it holds that $\epsilon \leq 1/6$ when $T$ is sufficiently large.} %\bo{please change $\log_2^{3/2}T$ to $(\log_2 T)^{3/2}$ and change $(\log_2 T)^3$ to $(\log_2 T)^3$ throughout the proof}

In the second case of $\sqrt{\Bex} > c_1 K^{1/6}T^{1/3}$, we have
\begin{align}
\ex{R_T} &\overset{\text{(a)}}{\geq} \frac{\epsilon T}{3} - \frac{81\ln 2 \cdot \epsilon^4 T^2  (\log_2 T)^3}{K } \cdot \frac{ \sqrt{\Bex}}{c_1 K^{1/6}T^{1/3}}  - \frac{ 18\sqrt{\ln 2} \cdot \epsilon^2 T (\log_2 T)^{3/2}}{\sqrt{K}}  \sqrt{\Bex}\nonumber\\
&= \frac{\epsilon T}{3} - \frac{81 \ln 2 \cdot \epsilon^4 T^{5/3}  (\log_2 T)^3}{c_1 K^{7/6} } \sqrt{\Bex} - \frac{ 18\sqrt{\ln 2} \cdot \epsilon^2 T (\log_2 T)^{3/2}}{\sqrt{K}}  \sqrt{\Bex}\nonumber\\
&\overset{\text{(b)}}{=} \frac{c_3 T\sqrt{K}}{3 (\log_2 T)^{3/2} \cdot \sqrt{\Bex}} - \frac{81\ln 2\cdot c_3^4 K^{5/6} T^{5/3}}{c_1(\log_2 T)^3 \cdot (\Bex)^{3/2}} - \frac{ 18\sqrt{\ln2}\cdot c_3^2\sqrt{K} T }{(\log_2 T)^{3/2} \cdot \sqrt{\Bex}}\nonumber\\
&\overset{\text{(c)}}{\geq} \frac{c_3 T\sqrt{K}}{3 (\log_2 T)^{3/2}\cdot \sqrt{\Bex}} - \frac{81 \ln 2 \cdot c_3^4 \sqrt{K} T }{c_1^3 (\log_2 T)^3 \cdot \sqrt{\Bex}} - \frac{18\sqrt{\ln 2} \cdot c_3^2 \sqrt{K} T }{(\log_2 T)^{3/2} \cdot \sqrt{\Bex}}\nonumber \\
&\geq \left( \frac{c_3}{3} - \frac{81\ln 2 \cdot c_3^4}{c_1^3} - 18\sqrt{\ln 2}\cdot c_3^2 \right)\frac{\sqrt{K} T }{(\log_2 T)^3 \cdot \sqrt{\Bex}}\nonumber\\
&= \widetilde{\Omega}\left(T\sqrt{K/\Bex}\right)\label{eqn:LB_case_2},
\end{align}
where (a) is due to $\frac{\sqrt{\Bex}}{c_1 K^{1/6}T^{1/3}}  > 1$, 
%\bo{in the second step, shouldn't it be $T^{5/3}$ instead of $T^{5/6}$?}\duo{after my calculation, seems that you are correct} 
(b) is obtained by choosing $\epsilon =\frac{c_3\sqrt{K}}{(\log_2 T)^{3/2} \sqrt{\Bex}}$ (where $c_3>0$ satisfies $ {1}/{3} - 81\ln 2 \cdot {c_3^3}/c_1^3 - 18\sqrt{\ln 2}\cdot c_3 > 0$), and (c) again is due to $\frac{\sqrt{\Bex}}{c_1 K^{1/6}T^{1/3}}  > 1$ (applied to the second term).
Since $K\leq T$, we have $\epsilon =\frac{c_3\sqrt{K}}{(\log_2 T)^{3/2} \sqrt{\Bex}}\leq \frac{c_3 K^{1/3}}{c_1 T^{1/3}(\log_2 T)^{3/2}} \le 1/6$ when $T$ is sufficiently large.

%Now, we can conclude that in both cases, for sufficiently large $T$, we indeed have $\epsilon \le 1/6$. \bo{you need to show the same for the first case. or you can explicitly say that in both cases, this holds for sufficiently large T}\duo{fixed}

%\bo{the following paragraph is unclear and needs to be rewritten}\duo{starting from here, I re-write until the end.}

Now, we want to relax the assumption that the deterministic learning algorithm makes no more than $\epsilon T$ switches. Similar to the proof of Theorem~2 in \citet{dekel2014bandits}, we consider the following: If the learning algorithm makes more than $\epsilon T$ switches, then we halt the algorithm at the point when there are exactly $\epsilon T$ switches and repeat its action after the last switch throughout the rest of the game. We use $R_T^{\text{halt}}$ to denote the regret of this halted algorithm over the same loss sequence as in $R_T$.

We consider two cases for the number of switches made by the original learning algorithm: $\bar{S}\leq \epsilon T$ and $\bar{S} > \epsilon T$. 

In the first case of $\bar{S}\leq \epsilon T$, halting does not happen, and trivially, we have $R_T^{\text{halt}}=R_T$. 

In the second case of $\bar{S} > \epsilon T$, the original learning algorithm makes more than $\epsilon T$ switches. We use $T'$ to denote the round index at which the $\epsilon T$-th switch happens. Clearly, we have $\epsilon T \leq T'$. As a result, the halted algorithm keeps playing action $X_{T'}$ from round $T'+1$ to the end of the game. We now rewrite $R_T^{\text{halt}}$ and $R_T$ as follows:
\begin{align}
R_T &=  \sum_{t=1}^{T'}\left(\ell_{t}[X_{t}] - \ell_{t}[k^*] \right) + \epsilon T + \sum_{t=T'+1}^{T}\left(\ell_{t}[X_{t}] - \ell_{t}[k^*] \right) + (\Bar{S} - \epsilon T), \nonumber \\
R_T^{\text{halt}} &=  \sum_{t=1}^{T'}\left(\ell_{t}[X_{t}] - \ell_{t}[k^*] \right) + \epsilon T + \sum_{t=T'+1}^{T}\left(\ell_{t}[X_{T'}] - \ell_{t}[k^*] \right). \nonumber
\end{align}
By taking the difference between $R_T^{\text{halt}}$ and $R_T$ and then taking the expectation with respect to the randomness from loss generation, we have
$$\ex{R_T^{\text{halt}} - R_T} = \underbrace{\ex{\sum_{t=T'+1}^T \left(\ell_{t}[X_{T'}] - \ell_{t}[X_t]\right)}}_{\le \epsilon T} + \underbrace{\ex{\epsilon T - \Bar{S}}}_{<0} \leq \epsilon T.
$$
To see this, we first observe that at each round, the loss gap between actions is either $\epsilon$ or 0 \emph{in expectation} (because the Gaussian noise we add has zero mean). Therefore, the first term $\ex{\sum_{t=T'+1}^T \left(\ell_{t}[X_{T'}] - \ell_{t}[X_t]\right)}$ can be bounded by $\epsilon T$ (i.e., the gap at each round multiplied by the time horizon).
Since the original learning algorithm makes more than $\epsilon T$ switches, we have
$\ex{R_T}\geq \epsilon T$, which implies $\ex{R_T^{\text{halt}}}\leq 2\ex{R_T}$.

Combining the above two cases yields $\ex{R_T^{\text{halt}}}\leq 2\ex{R_T}$. Since we already obtain a lower bound for $\ex{R_T^{\text{halt}}}$ (i.e., Eqs.~(\ref{eqn:LB_case_1}) and (\ref{eqn:LB_case_2})), the same lower bound also holds for $\ex{R_T}$ (within a constant factor of 2). 

Finally, we complete the proof by applying Yao's principle.
\end{proof}

We also give two remarks about technical details in the proof of Proposition~\ref{prop:bandit_extra_lower_bound}.

\begin{remark}
    To conclude that $\ex{R_T^{\text{halt}}}\leq 2\ex{R_T}$, \citet{dekel2014bandits} shows that $R_T^{\text{halt}}\leq R_T + \epsilon T \leq 2 R_T$, which is a stronger result compared to what is needed in an expected sense. This stronger result relies on the fact that the loss gap between actions is either $\epsilon$ or 0 at each round. However, this may not be true anymore after introducing an additional action-dependent noise as in \citet{shi2022power}. Despite this difference, one can still show $\ex{R_T^{\text{halt}}}\leq \ex{R_T} + \epsilon T \leq 2\ex{R_T}$, which is used to prove Proposition~\ref{prop:bandit_extra_lower_bound}.
    %, which is sufficient, without the need for a stronger and unnecessary condition as in \citet{dekel2014bandits}.
\end{remark}

\begin{remark}
    Some readers may ask: If the deterministic algorithm switches more than $\epsilon T$ times, should not the switching costs already imply the desired lower bound on the regret? Why is it necessary to show a reduction from switch-limited algorithms to arbitrary algorithms? To see this, we note that the lower bound of $\wtOmega(T^{2/3})$ is obtained after selecting $\epsilon $ to be of order $\wtTheta(T^{-1/3})$, while such selection is based on the previous analysis, which is further based on the assumption that the algorithm makes \emph{no more than} $\epsilon T$ switches. Therefore, the reduction from switch-limited algorithms to arbitrary algorithms is necessary.
    %this is not a feasible option due to the contradiction.
\end{remark}

\section{Proof of Proposition \ref{prop:uniform_alg_upper_bound}}\label{app:uniform_alg_upper_bound_proof}

%\bo{If these lemmas are from another paper, why do you need to prove them again? If they are new, although similar to those in the literature, you shouldn't just call them lemmas from other papers.}\duo{They are very simple batched-version extensions but not exactly the same as in the original papers, so I prove again, though there's nothing new in the proof.}

Before proving Proposition \ref{prop:uniform_alg_upper_bound}, we first present two lemmas on the properties of SD. These are straightforward extensions of Lemmas~1 and 2 in \citet{geulen2010regret} to their batched versions.
A key difference is that we consider batches instead of rounds. While the proofs follow the same line of analysis, we provide the proofs below for completeness.
\begin{lemma}%\citep[Lemma~1]{geulen2010regret}
\label{lem:SD_regret}
For the instantiations $\pi_{\text{full}}$ and $\pi_{\text{flex}}$ of algorithm~\ref{alg:framework}, over any loss sequence, we have $$\bP\left( A_b = k\right) = \bE[w_b[k]], ~\forall k\in [K] , \forall b \in [N].$$
\end{lemma}

\begin{proof}[Proof of \lemref{SD_regret}]
We first note that in these two instantiations of \algref{framework}, the feedback depends on the randomly chosen $u_b$ only, and is thus independent of what actions are taken by the learner throughout the game. As a result, the whole feedback sequence $\widehat{\ell}_{1:N}$ can be fixed even before the learner's actions are determined.

In the following, we will show by induction that conditioning on any random feedback sequence $\widehat{\ell}_{1:N}$, it holds (almost surely) that $\bP\left( A_b = k|\widehat{\ell}_{1:N}\right) = w_b[k], \forall k\in[K], \forall b\in [N] $.
%\xingyu{we can try to add $\omega$ into $A_b$ and $w_b$, something like $A_b^{\omega}$ and $w_b^{\omega}$. Then, for each realization of $\widehat{\ell}$, we basically say for any $\omega$.}\duo{I'm not familiar with such notation. Any reference on this?}

% \xingyu{More on the above Lemma 1: It seems to me that if we have Lemma~\ref{lem:SD_regret} as stated above (i.e., expectation on the RHS), we may not be able to establish the required result. The key is to show that $\ex{\ell_i(A_b))} = \ex{\inner{w_b}{\ell_i}}$. Then, Let first see what is corresponding part in SD paper. In particular, $\ex{\ell_i(A_b)} = \sum_k \prob{A_b = k} \ell_i[k] = \inner{w_b}{\ell_i}$, where we have used $\prob{A_b = k} = w_b$ in the SD paper. Then, we use the OMD bound for the term $\inner{w_b}{\ell_i}$.

% Now, for our case, we have 
% \begin{align}
%     \ex{\ell_i(A_b)} = \ex{\ex{\ell_i(A_b)| \widehat{\ell}_{b\ge 1}}} = \ex{ \sum_k \prob{A_b = k | \widehat{\ell}_{b\ge 1}}\ell_i(k) }
% \end{align}
% } 

The base case of $b=1$ is trivial due to the algorithm design (specifically, the uniform action weight initialization). In the following, we move forward to the induction step, i.e., we show that for every $b\geq 2$, if it holds that $\bP\left( A_{b-1} = k | \widehat{\ell}_{1:N} \right) = w_{b-1}[k], ~\forall k \in [K]$, then it also holds that $\bP\left( A_b = k|\widehat{\ell}_{1:N} \right) = w_b[k], ~\forall k \in [K]$, as follows:

\begin{align}
    \bP\left( A_b = k | \widehat{\ell}_{1:N} \right)&\overset{\text{(a)}}{=} W_b[k]/ W_{b-1}[k] \cdot \bP\left( A_{b-1} = k | \widehat{\ell}_{1:N} \right) \nonumber \\
    &\quad + \sum_{i=1}^K \left( 1-W_b[i]/ W_{b-1}[i] \right) \cdot w_b[k] \cdot \bP\left( A_{b-1} = i | \widehat{\ell}_{1:N} \right) \nonumber \\
    &\overset{\text{(b)}}{=}W_b[k]/ W_{b-1}[k] \cdot w_{b-1}[k] + w_b[k] \cdot \sum_{i=1}^K \left( 1-W_b[i]/ W_{b-1}[i] \right) \cdot w_{b-1}[i]  \nonumber \\
    &=\frac{W_b[k]}{W_{b-1}[k]}  \cdot \frac{W_{b-1}[k]}{\sum_{i=1}^K {W_{b-1}[i]}} + w_b[k] \cdot \sum_{i=1}^K \left( \frac{W_{b-1}[i] - W_b[i]}{W_{b-1}[i]} \right) \cdot \frac{W_{b-1}[i]}{\sum_{j=1}^K {W_{b-1}[j]}}\nonumber \\
    &=\frac{W_b[k]}{\sum_{i=1}^K {W_{b-1}[i]}} + w_b[k] \cdot  \frac{\sum_{i=1}^K {W_{b-1}[i]} - \sum_{i=1}^K {W_{b}[i]}}{\sum_{i=1}^K {W_{b-1}[i]}}\nonumber \\
    &=w_b[k] \cdot \frac{\sum_{i=1}^K {W_{b}[i]}}{\sum_{i=1}^K {W_{b-1}[i]}} + w_b[k] \cdot \left(1-\frac{\sum_{i=1}^K {W_b[i]}}{\sum_{i=1}^K {W_{b-1}[i]}}\right)\nonumber \\
    &=w_b[k],\nonumber
\end{align}
where (a) is due to Line \ref{line:next_decision} in Algorithm \ref{alg:framework} (specifically, there are two disjoint cases for action $k$ to be played in batch $b$: (i) action $k$ was played in batch $b-1$, and it does not change according to probability $\exp(-\eta \cdot \widehat{\ell}_b[k]) = {W_b[k]}/{W_{b-1}[k]}$; (ii) action $k$ is selected based on fresh randomness (i.e., the previous action $i$ does not stay unchanged with probability $(1 - {W_b[i]}/{W_{b-1}[i]})$), regardless of which action is played in batch $b-1$), and (b) is from the inductive hypothesis that $\bP\left( A_{b-1} = i|\widehat{\ell}_{1:N}  \right) = w_b[i], \forall i \in [K]$.

Finally, taking the expectation on both sides of the above completes the proof.
\end{proof}

Lemma~\ref{lem:SD_regret} implies that SD does not change the marginal distribution of action selection. Hence, one still enjoys the same regret guarantee as that of standard OMD (i.e., without SD).
\begin{lemma}%\citep[Lemma~2]{geulen2010regret}
\label{lem:SD_switch}
%Running \algref{framework} as specified in Lemma~\ref{lem:SD_regret}, 
For the instantiations $\pi_{\text{full}}$ and $\pi_{\text{flex}}$ of algorithm~\ref{alg:framework}, over any loss sequence, the expected number of switches satisfies the following: 
$$\bE \left[\sum_{t=1}^T \bI\{X_{t} \neq X_{t-1}\} \right]\leq \sum_{b=2}^{N}\eta\cdot \ex{\widehat{\ell}_{b-1}[A_{b-1}]}.$$
\end{lemma}
\begin{proof}[Proof of \lemref{SD_switch}]
Based on the definition of switching costs, we have
\begin{align}
\bE \left[\sum_{t=1}^T \bI_{\{X_{t} \neq X_{t-1}\}} \right] &\overset{\text{(a)}}{=} \bE \left[ \sum_{b=2}^{N} \bI_{\{A_{b} \neq A_{b-1}\}} \right]\nonumber\\
&= \sum_{b=2}^{N} \bE \left[ \bE \left[ \bI_{\{A_{b} \neq A_{b-1}\}} |\widehat{\ell}_{1:b-1} , A_{b-1}  \right] \right]\nonumber\\
&\overset{\text{(b)}}{\leq}  \sum_{b=2}^{N}  \bE \left[1 - \exp(-\eta \cdot \widehat{\ell}_{b-1}[A_{b-1}]) \right] \nonumber\\
&\overset{\text{(c)}}{\leq}  \sum_{b=2}^{N}\eta\cdot \ex{\widehat{\ell}_{b-1}[A_{b-1}]}, \nonumber
\end{align}
where (a) is because switching happens only between two consecutive batches, (b) is due to the action selection rule of Algorithm \ref{alg:framework} (Line~\ref{line:next_decision}), and (c) is from elementary inequality: $1-\exp(-x) \leq x, \forall x>0$.
\end{proof}
\begin{proof}[Proof of Proposition \ref{prop:uniform_alg_upper_bound}]

Our goal here is to establish an upper bound on the expected regret for algorithm $R_T^{\pi_{\text{full}}}$, which consists of two parts (cf.~Eqs.~\eqref{eq:reg-def} and~\eqref{eqn:reg_def}): (i) standard regret in terms of loss and (ii) switching cost. We will establish bounds on both respectively, hence obtaining the final result in the proposition.

To start with, let us establish an upper bound on the standard regret in terms of loss. To this end, we will build upon the classic analysis of OMD (cf.~\citep[Section 6.6]{orabona2019modern}). That is, for any (random) sequence $\widehat{\ell}_{1:N}\in [0, \infty)^{K}$, learning rate $\eta>0$, and vector $u$ such that its $Y_u$-th coordinate is 1 and all the others are 0, it holds almost surely that
\begin{align}
     \sum_{b=1}^N \left\langle w_b - u, \widehat{\ell}_b \right\rangle &\leq \frac{\ln K}{\eta} + \frac{\eta}{2} \sum_{b=1}^N \sum_{k=1}^K (\widehat{\ell}_b[k])^2 \cdot w_b[k],\nonumber
\end{align}
where $w_{b}[k] = \frac{W_{b}[k]}{\sum_{i=1}^{K}W_{b}[i]}$ and $W_{b}[k] = W_{b-1}[k] \cdot \exp(-\eta \cdot \widehat{\ell}_{b-1}[k]), \forall k\in[K]$, i.e., line~\ref{line:alg1_OMD_update} in Algorithm~\ref{alg:framework}. Taking the expectation on both sides yields that 
\begin{align}
    \bE\left[ \sum_{b=1}^N \left\langle w_b - u, \widehat{\ell}_b \right\rangle	 \right] &\leq \frac{\ln K}{\eta} + \frac{\eta}{2} \sum_{b=1}^N \sum_{k=1}^K \bE \left[ (\widehat{\ell}_b[k])^2 \cdot w_b[k] \right]\nonumber \\
    &= \frac{\ln K}{\eta} + \frac{\eta}{2} \sum_{b=1}^N \sum_{k=1}^K \bE \left[ w_b[k] \cdot \bE \left[ (\widehat{\ell}_b[k])^2 | \widehat{\ell}_{1:b-1}  \right] \right]\nonumber \\
    &\overset{\text{(a)}}{=} \frac{\ln K}{\eta} + \frac{\eta}{2} \sum_{b=1}^N \sum_{k=1}^K \bE \left[  w_b[k] \cdot  \left(\ell_{u_b}[k]\right)^2 \right]\nonumber \\
    &\overset{\text{(b)}}{\leq} \frac{\ln K}{\eta} + \frac{\eta B}{2K},\label{eqn:generel_OMD_regret_full}
\end{align}
where (a) follows from the algorithm design of $\pi_{\text{full}}$, i.e., $\widehat{\ell}_b = \ell_{u_b}$ and $u_b$ is a randomly selected time slot within batch $b$, and (b) comes from the boundedness of losses, the fact that $\sum_{k=1}^K w_b[k] = 1, \forall b\in\left[N\right]$, and noting that $N = B/K$.

We can further rewrite the left-hand-side (LHS) of the above inequality as follows:
\begin{align}
\mathbb{E}\left[ \sum_{b=1}^{N} \left\langle w_b - u, \widehat{\ell}_b \right\rangle	 \right] &= \sum_{b=1}^{N} \mathbb{E}\left[ \mathbb{E}\left[ \left\langle w_b - u, \widehat{\ell}_b \right\rangle |\widehat{\ell}_{1:b-1}  \right] \right]\nonumber \\
&= \sum_{b=1}^{N} \mathbb{E}\left[  \left\langle w_b - u, \mathbb{E}\left[ \widehat{\ell}_b | \widehat{\ell}_{1:b-1} \right] \right\rangle  \right]\nonumber \\
&\overset{\text{(a)}}{=} \sum_{b=1}^{N} \mathbb{E}\left[  \left\langle w_b - u, \frac{ \sum_{t = (b-1)\tau+1}^{b\tau} \ell_{t} }{\tau} \right\rangle  \right] \nonumber\\
&= \frac{1}{\tau} \sum_{b=1}^{N} \sum_{t = (b-1)\tau+1}^{b\tau} \mathbb{E}\left[ \ell_t[A_b] - \ell_t[Y_u]  \right]\nonumber \\
&= \frac{1}{\tau} \sum_{t=1}^T \mathbb{E}\left[  \ell_t[X_t] - \ell_t[Y_u]  \right],\label{eqn:batch_reduction_full}
\end{align}
where (a) holds since for any $k \in [K]$, we have $\mathbb{E} \left[ \widehat{\ell}_b[k] |\widehat{\ell}_{1:b-1} \right] = \sum_{t = (b-1)\tau+1}^{b\tau}   \ell_{t}[k]/\tau$. This is true since under $\pi_{\text{full}}$, the choice of $u_b$ for constructing $\widehat{\ell}_b$ is a round index chosen uniformly at random from the current batch $b$. 

Now, replacing the LHS of Eq.~\eqref{eqn:generel_OMD_regret_full}
with Eq.~\eqref{eqn:batch_reduction_full}, yields 
\begin{align}
    \sum_{t=1}^T \mathbb{E}\left[  \ell_t[X_t] - \ell_t[Y_u]\right] &\leq \tau \cdot \left( \frac{\ln K}{\eta} + \frac{\eta B}{2K}\right)\nonumber\\
    &=\frac{KT\ln K}{\eta B} + \frac{\eta T}{2 },\label{eqn:last_step_regret_full}
\end{align}
where the last step is due to $\tau=\frac{KT}{B}$.

After obtaining the above upper bound on the standard regret (i.e., without switching costs), we now turn to bound the switching costs under $\pi_{\text{full}}$. To this end, we directly leverage Lemma~\ref{lem:SD_switch} along with the loss estimate $\widehat{\ell}_b = \ell_{u_b}$ to obtain that 
\begin{align}
    \bE \left[\sum_{t=1}^T \bI_{\{X_{t} \neq X_{t-1}\}} \right] \leq \sum_{b=2}^{N}\eta\cdot \ex{\widehat{\ell}_{b-1}[A_{b-1}]} = \sum_{b=2}^{B/K}\eta\cdot \ex{\widehat{\ell}_{b-1}[A_{b-1}]} \leq \frac{\eta B}{K}.  \label{eqn:switching_cost_upper_bound_full}
\end{align}

Finally, combining Eqs.~(\ref{eqn:last_step_regret_full}) and (\ref{eqn:switching_cost_upper_bound_full}), we can bound the total regret as follows:
\begin{align*}
   R_T^{\pi_{\mathrm{full}}} &\leq \frac{KT\ln K}{\eta B} + \frac{\eta T}{2} + \frac{\eta B}{K}\\
    &\overset{\text{(a)}}{\leq}  \frac{KT\ln K}{\eta B} + \frac{3\eta T}{2}\\
    &\overset{\text{(b)}}{=}  T\sqrt{\frac{6K\ln K}{B}},
\end{align*}
where (a) is from $B \le KT$, and (b) is obtained by choosing $\eta = \sqrt{\frac{2K\ln K}{3B}}$. Hence, we have completed the proof of Proposition~\ref{prop:uniform_alg_upper_bound}.
% \textcolor{blue}{At this point, we have obtained an upper bound on the regret without switching costs. Now, we turn to bound the switching costs in the following.} In particular, according to Lemma \ref{lem:SD_switch}, we have
% \begin{align}
%     \bE \left[\sum_{t=1}^T \bI_{\{X_{t} \neq X_{t-1}\}} \right] \leq \frac{\eta B}{K}.  \label{eqn:switching_cost_upper_bound_full}
% \end{align}
% Combining Eqs.~(\ref{eqn:last_step_regret_full}) and (\ref{eqn:switching_cost_upper_bound_full}), we can bound the total regret as follows:
% \begin{align*}
%    R_T^{\pi_{\mathrm{full}}} &\leq \frac{KT\ln K}{\eta B} + \frac{\eta T}{2} + \frac{\eta B}{K}\\
%     &\overset{\text{(a)}}{\leq}  \frac{KT\ln K}{\eta B} + \frac{3\eta T}{2}\\
%     &\overset{\text{(b)}}{=}  T\sqrt{\frac{6K\ln K}{B}},
% \end{align*}
% where (a) is from $B \le KT$, and (b) is obtained by choosing $\eta = \sqrt{\frac{2K\ln K}{3B}}$.
\end{proof}

\section{Proof of Proposition \ref{prop:flex upper bound} }\label{app:B<T_upper_bound_proof}
\begin{proof}[Proof of Proposition \ref{prop:flex upper bound}]

The organization of this proof is the same as that of Proposition~\ref{prop:uniform_alg_upper_bound}, and the only difference lies in the loss estimate in this instantiation.

We first consider the case when $B<T$, i.e., $M$ can be any integer from $[K]$. Recall that $B$ is the total observation budget and $M$ is the number of observations made in each batch.
%\xingyu{It might be better to give the name for $M$, i.e., the number of observations, to remind reader about the meaning of each variable.}\duo{added}
Similarly, we have, for any (random) sequence $\widehat{\ell}_1,\dots,\widehat{\ell}_N \in [0, \infty)^{K}$, learning rate $\eta>0$, and vector $u$ such that its $Y_u$-th coordinate is 1 and all the others are 0, it holds almost surely that
\begin{align}
     \sum_{b=1}^N \left\langle w_b - u, \widehat{\ell}_b \right\rangle &\leq \frac{\ln K}{\eta} + \frac{\eta}{2} \sum_{b=1}^N \sum_{k=1}^K (\widehat{\ell}_b[k])^2 \cdot w_b[k].\nonumber
\end{align}
Taking the expectation on both sides yields that 
\begin{align}
    \bE\left[ \sum_{b=1}^N \left\langle w_b - u, \widehat{\ell}_b \right\rangle	 \right] &\leq \frac{\ln K}{\eta} + \frac{\eta}{2} \sum_{b=1}^N \sum_{k=1}^K \bE \left[ (\widehat{\ell}_b[k])^2 \cdot w_b[k] \right]\nonumber \\
    &= \frac{\ln K}{\eta} + \frac{\eta}{2} \sum_{b=1}^N \sum_{k=1}^K \bE \left[ \bE \left[ (\widehat{\ell}_b[k])^2 \cdot w_b[k] \bigg| \widehat{\ell}_{1:b-1} \right] \right]\nonumber  \\
    &\overset{\text{(a)}}{=} \frac{\ln K}{\eta} + \frac{\eta}{2} \sum_{b=1}^N \sum_{k=1}^K \bE \left[ \left(\frac{\ell_{u_b}[k]}{M/K}\right)^2 \cdot w_b[k] \cdot \frac{M}{K} + 0 \cdot (1-\frac{M}{K}) \right]\nonumber  \\
    &\overset{\text{(b)}}{\leq} \frac{\ln K}{\eta} + \frac{\eta  KB}{2M^2},\label{eqn:generel_OMD_regret_generel_full}
\end{align}
where (a) follows from the algorithm design of $\pi_{\text{flex}}$, i.e., the loss estimate $\widehat{\ell}_b[k] = \bI\{k\in \mathcal{O}_{u_b}\} \cdot \frac{\ell_{u_b}[k]}{M/K}$ and $u_b$ is a randomly selected time slot within batch $b$, and (b) comes from the boundedness of losses, the fact that $\sum_{k=1}^K w_b[k] = 1, \forall b\in\left[N\right]$, and noting that $N = B/K$.

We can further rewrite the LHS of the above inequality as follows:
\begin{align}
\mathbb{E}\left[ \sum_{b=1}^{N} \left\langle w_b - u, \widehat{\ell}_b \right\rangle	 \right] &= \sum_{b=1}^{N} \mathbb{E}\left[ \mathbb{E}\left[ \left\langle w_b - u, \widehat{\ell}_b \right\rangle \bigg| \widehat{\ell}_{1:b-1} \right] \right]\nonumber \\
&= \sum_{b=1}^{N} \mathbb{E}\left[  \left\langle w_b - u, \mathbb{E}\left[ \widehat{\ell}_b \bigg| \widehat{\ell}_{1:b-1} \right] \right\rangle  \right]\nonumber \\
&\overset{\text{(a)}}{=} \sum_{b=1}^{N} \mathbb{E}\left[  \left\langle w_b - u, \frac{ \sum_{t = (b-1)\tau+1}^{b\tau} \ell_{t} }{\tau} \right\rangle  \right] \nonumber\\
&= \frac{1}{\tau} \sum_{b=1}^{N} \sum_{t = (b-1)\tau+1}^{b\tau} \mathbb{E}\left[ \ell_t[A_b] - \ell_t[Y_u]  \right]\nonumber \\
&= \frac{1}{\tau} \sum_{t=1}^T \mathbb{E}\left[  \ell_t[X_t] - \ell_t[Y_u] \right], \label{eqn:batch_reduction_general_full}
\end{align}
where (a) holds since for any $k\in [K]$, we have $\mathbb{E} \left[ \widehat{\ell}_b[k] \bigg|\widehat{\ell}_{1:b-1} \right] = (1-\frac{M}{K}) \cdot 0 + \frac{M}{K} \cdot \sum_{t = (b-1)\tau+1}^{b\tau} \frac{1}{\tau} \cdot  \frac{\ell_{t}[k]}{M/K} = { \sum_{t = (b-1)\tau+1}^{b\tau} \ell_{t}[k] }/{\tau}$. This is true since under $\pi_{\text{flex}}$, the choices of both $u_b$ and $\mathcal{O}_t$ for constructing $\widehat{\ell}_b$ are uniformly random and independent with each other.

Now, replacing the LHS of Eqs. (\ref{eqn:generel_OMD_regret_generel_full}) with (\ref{eqn:batch_reduction_general_full}), yields
\begin{align}
    \sum_{t=1}^T \mathbb{E}\left[  \ell_t[X_t] - \ell_t[Y_u]\right] &\leq \tau\cdot\left( \frac{\ln K}{\eta} + \frac{\eta  KB}{2M^2}\right)\nonumber\\
    &=\frac{MT\ln K}{\eta B} + \frac{\eta T K}{2M },\label{eqn:last_step_regret_general_full}
\end{align}
where in the equality, we replace $\tau$ with $\frac{MT}{B}$.

After obtaining the above upper bound on the standard regret (i.e., without switching cost), we now turn to bound the switching costs under $\pi_{\text{flex}}$. To this end, we directly leverage Lemma~\ref{lem:SD_switch} along with the loss estimate $\widehat{\ell}_b[k] = \bI\{k\in \mathcal{O}_{u_b}\} \cdot \frac{\ell_{u_b}[k]}{M/K}$ to obtain that
\begin{align}
    \bE \left[\sum_{t=1}^T \bI_{\{X_{t} \neq X_{t-1}\}} \right] \leq \sum_{b=2}^{N}\eta\cdot \ex{\widehat{\ell}_{b-1}[A_{b-1}]} = \sum_{b=2}^{B/M}\eta\cdot \ex{\widehat{\ell}_{b-1}[A_{b-1}]} \leq \eta \frac{B}{M} \cdot \frac{K}{M}=\frac{\eta BK}{M^2}.  \label{eqn:switching_cost_upper_bound_general_full}
\end{align}
Finally, combining Eqs. (\ref{eqn:last_step_regret_general_full}) and (\ref{eqn:switching_cost_upper_bound_general_full}), we can bound the total regret as follows:
\begin{align}
   R_T^{\pi_{\mathrm{flex}}} &\leq \frac{MT\ln K}{\eta B} + \frac{\eta T K}{2M} + \frac{\eta BK}{M^2}\nonumber\\
    &\overset{\text{(a)}}{\leq}  \frac{MT\ln K}{\eta B} + \frac{3\eta KT}{2M}\nonumber\\
    &\overset{\text{(b)}}{=} T\sqrt{\frac{6K\ln K}{B}}\nonumber,
\end{align}
where (a) is from $B < MT$ (recall that now we have $B<T$ and $M\geq 1$), and (b) is obtained by choosing $\eta = M\sqrt{\frac{2\ln K}{3BK}}$. Hence, we have completed the proof of Proposition \ref{prop:flex upper bound}.

The proof for the case of $B\geq T$ is exactly the same, except for the (implicit) fact that we need batch size $\tau$ to be well-defined, i.e., $\tau\geq 1$. That is why in this case $M$ is less flexible: now $M$ needs to be sufficiently large to fully exploit the total budget.
\end{proof}

\section{Proof of Proposition \ref{prop:bandit_upper_bound}}\label{app:bandit_upper_bound_proof}
\begin{proof}[Proof of Proposition \ref{prop:bandit_upper_bound}]

The organization of this proof is the same as that of Proposition~\ref{prop:uniform_alg_upper_bound}, and the main difference lies in the commonly-used importance-weighted estimator for bandit feedback. In addition, we note that it is now sufficient to directly bounding the switching costs by the number of batches.

We still start with the same fundamental conclusion in OMD analysis. Specifically, for any (random) sequence $\widehat{\ell}_1,\dots,\widehat{\ell}_N \in [0, \infty)^{K}$, learning rate $\eta>0$, and vector $u$ such that its $Y_u$-th coordinate is 1 and all the others are 0, it holds almost surely that
\begin{equation}
 \sum_{b=1}^{N} \left\langle w_b - u, \widehat{\ell}_b \right\rangle  \leq \frac{\ln K}{\eta} + \frac{\eta}{2}\sum_{b=1}^{N} \sum_{k=1}^K (\widehat{\ell}_b[k])^2 w_b[k].\nonumber \label{eqn:5_1}
\end{equation}

Taking the expectation on both sides, we have
\begin{align}
\bE\left[\sum_{b=1}^{N} \left\langle w_b - u, \widehat{\ell}_b \right\rangle \right] &\leq \frac{\ln K}{\eta} + \frac{\eta}{2}\sum_{b=1}^{N} \bE\left[\sum_{k=1}^K (\widehat{\ell}_b[k])^2 w_b[k]\right]\nonumber\\
&= \frac{\ln K}{\eta} + \frac{\eta}{2}\sum_{b=1}^{N} \ex{\ex{\sum_{k=1}^K (\widehat{\ell}_b[k])^2 w_b[k]\bigg|\widehat{\ell}_{1:b-1}}}\nonumber\\
&\overset{\text{(a)}}{=} \frac{\ln K}{\eta} + \frac{\eta}{2}\sum_{b=1}^{N}\ex{ \sum_{k=1}^K w_b[k] \cdot \frac{1}{\tau} \cdot \sum_{t = (b-1)\tau+1}^{b\tau} \frac{(\ell_t[k])^2}{(w_b[k])^2} \cdot w_b[k]}\nonumber\\
&\overset{\text{(b)}}{\leq} \frac{\ln K}{\eta} + \frac{\eta}{2}NK\nonumber\\
&\overset{\text{(c)}}{=} \sqrt{2NK\ln K},\label{eqn:batch_bandit_estimate_loss}
\end{align}
where (a) follows from the algorithm design of $\pi_{\text{b}}$, i.e., the loss estimate $\widehat{\ell}_b[k] = \bI\{k\in \mathcal{O}_{u_b}\} \cdot \frac{\ell_{u_b}[k]}{w_b[k]} $ for all $k\in [K]$, (b) follows from the assumption that all losses are bounded by one and the fact that $\sum_{k=1}^K w_b[k] = 1, \forall b \in [N]$, and (c) is obtained by choosing $\eta = \sqrt{\frac{2\ln K}{NK}}$.
%\duo{I still think that there shouldn't be an expectation at the LHS above, since here we fix the loss sequence, i.e., all $u_b$'s, in this case, there's no random variable at LHS, am I wrong?} \xingyu{I think Eq. 23 is very confusing to me (why? since in standard result of EXP3, we do not have $w_b$ on LHS, which is a probability vector. Rather, we only has the actual pulled action. Then, we need expectation). A suggestion is to directly use OMD with estimations to prove regret of Algorithm 1 as discussed above (nearly the same as in the non-batch case .), i.e., follow Thm 12.5 in my note \url{https://drive.google.com/file/d/1k-yWao8HcYjbWYdpcMb-SzQvRbiNrAd1/view}}\duo{sure. I'll complete the proof, but what I mean by eqn.23 (removing the expectation and before selecting $\eta$) is the eqn 12.16 in your note, if that helps to clarify}
%\xingyu{yes, it is equal to 12.16. Then, you just need to bound the local norms, then done.}
%\duo{I see, then seemingly we are on the same page :)}\xingyu{yes:) the main point is that it seems to me we cannot directly use reduction to EXP3 as in~\cite{Dekel2012OnlineBL}. Rather, we have to start from 12.16 to re-prove it.}

We can further rewrite the LHS of the above inequality as follows:
\begin{align}
\mathbb{E}\left[ \sum_{b=1}^{N} \left\langle w_b - u, \widehat{\ell}_b \right\rangle	 \right] &= \sum_{b=1}^{N} \mathbb{E}\left[ \mathbb{E}\left[ \left\langle w_b - u, \widehat{\ell}_b \right\rangle \bigg| \widehat{\ell}_{1:b-1} \right] \right]\nonumber \\
&= \sum_{b=1}^{N} \mathbb{E}\left[  \left\langle w_b - u, \mathbb{E}\left[ \widehat{\ell}_b \bigg| \widehat{\ell}_{1:b-1} \right] \right\rangle  \right]\nonumber \\
&\overset{\text{(a)}}{=} \sum_{b=1}^{N} \mathbb{E}\left[  \left\langle w_b - u, \frac{ \sum_{t = (b-1)\tau+1}^{b\tau} \ell_{t} }{\tau} \right\rangle  \right] \nonumber\\
&= \frac{1}{\tau} \sum_{b=1}^{N} \sum_{t = (b-1)\tau+1}^{b\tau} \mathbb{E}\left[ \ell_t[A_b] - \ell_t(Y_u)  \right]\nonumber \\
&= \frac{1}{\tau} \sum_{t=1}^T \mathbb{E}\left[  \ell_t[X_t] - \ell_t[Y_u]  \right] \label{eqn:batch_reduction_general_bandit},
\end{align}
where (a) holds since for any $k\in[K]$, we have $\mathbb{E} \left[ \widehat{\ell}_b[k] \bigg| \widehat{\ell}_{1:b-1} \right] = (1-w_b[k]) \cdot 0 + \sum_{t = (b-1)\tau+1}^{b\tau} \frac{1}{\tau} \cdot w_b[k] \cdot \frac{\ell_{t}[k]}{w_b[k]} = { \sum_{t = (b-1)\tau+1}^{b\tau} \ell_{t}[k] }/{\tau}$.

Now, replacing the LHS of Eq.~(\ref{eqn:batch_bandit_estimate_loss}) with Eq.~(\ref{eqn:batch_reduction_general_bandit}), yields
\begin{align}
\sum_{t=1}^T \mathbb{E}\left[  \ell_t[X_t] - \ell_t[Y_u]  \right] &\leq \tau \sqrt{2N K\ln K}\nonumber \\
&= \frac{T}{B} \cdot \sqrt{2 B K \ln K}\nonumber \\
&= O\left( T \sqrt{\frac{K\ln K}{B}}\right). \label{eqn:5_4}
\end{align}

When $B = O(\PTP)$, we have $\sum_{t=1}^T \mathbb{I}_{\{X_{t} \neq X_{t-1}\}} \leq B \leq O\left( T \sqrt{\frac{K\ln K}{B}}\right)$. Combining it with Eq.~(\ref{eqn:5_4}), we can conclude that both the standard regret and switching costs are of order $O\left( T \sqrt{\frac{K\ln K}{B}}\right)$, which gives us the desired result.
%\duo{typically we don't say $\leq O()$ right? we only have $=O()$ which in fact is $\in O()$, since $O()$ is a set of functions...}\xingyu{yes, it is a set. but we can say $\le$ and just use $\le$ here is fine}
%\xingyu{missing refs here.}
\end{proof}

\section{An Auxiliary Technical Result}\label{app:auxiliary}
% For the first auxiliary result, we show that when total budget $B\leq T$, using bandit feedback can match the lower bound of $\Omega(T\sqrt{K/B})$ under standard regret definition (i.e., without switching costs), which is formally stated in the following proposition.

In this section, we show that in the standard setting without switching costs, only using bandit feedback (e.g., algorithm $\pi_{\mathrm{b}}$) can also achieve optimal regret (i.e., matching the lower bound of $\Omega(T\sqrt{K/B})$, up to poly-logarithmic factors) in the full range of $B \in [K,T]$. We state this result in Proposition~\ref{prop:bandit_no_sc_reg}.

\begin{proposition}\label{prop:bandit_no_sc_reg}
%For any $B\in[K,T]$, the worst-case regret (without switching costs) under algorithm $\pi_{\mathrm{b}}$ is upper bounded by $ O(T\sqrt{K\ln K/B})$.
In the standard setting without switching costs,
for any $B\in[K,T]$, the worst-case regret under algorithm $\pi_{\mathrm{b}}$ is upper bounded by $R_T^{\pi_{\mathrm{b}}} = O(T\sqrt{K\ln K/B})$.
\end{proposition}
\begin{proof}[Proof of Proposition \ref{prop:bandit_no_sc_reg}]
    The proof follows the same line of analysis as that in the proof of Proposition~\ref{prop:bandit_upper_bound}, except that we only require $B<T$ (instead of $B=O(\PTP)$) and do not consider switching costs. Therefore, Eq. (\ref{eqn:5_4}) implies the following upper bound on the regret: $$\sum_{t=1}^T \mathbb{E}\left[  \ell_t[X_t] - \ell_t[Y_u]  \right] =  O\left( T \sqrt{K\ln K/B}\right).$$ Note that $Y_u$ can be any fixed action, including the best fixed action. This completes the proof.
\end{proof}